\documentclass[12pt]{article} 
\usepackage{amsmath,amsthm,amssymb,bm,mathrsfs}
\usepackage{graphicx,subfigure}
\usepackage{caption,enumerate,listings,setspace}
\usepackage[colorlinks,linkcolor=red,anchorcolor=blue,citecolor=blue]{hyperref}
\usepackage[margin=1in]{geometry}
\usepackage[title]{appendix}
\usepackage{enumitem}
\theoremstyle{plain}
\usepackage{booktabs}
\usepackage[numbers]{natbib}
\usepackage{array}
\usepackage{algorithm}
\usepackage{algpseudocode}
\usepackage{makecell}

\numberwithin{example}{section}

\usepackage[utf8x]{inputenc}
\usepackage{mathrsfs}
\usepackage{amssymb}
\usepackage{amsfonts}
\usepackage{amsmath}
\usepackage{amsthm,verbatim}
\usepackage{wrapfig}
\usepackage{multirow}
\usepackage{longtable}
\usepackage{enumerate}
\usepackage{color, graphicx}
\usepackage{tikz}

\def\boxit#1{\vbox{\hrule\hbox{\vrule\kern6pt
          \vbox{\kern6pt#1\kern6pt}\kern6pt\vrule}\hrule}}

\title{\textbf{
Downstream Task-Oriented Generative Model Selections on Synthetic Data Training for Fraud Detection Models}} 

\author
{
Yinan Cheng\thanks{Dept of Statistics, UC Davis. Email: ynccheng@ucdavis.edu},
Chi-Hua Wang \thanks{Department of Statistics and Data Science,  Email: chihuawang@ucla.edu},\\
Vamsi K. Potluru \thanks{J.P. Morgan AI Research}, Tucker Balch \thanks{J.P. Morgan AI Research},
Guang Cheng\thanks{Department of Statistics and Data Science. Email: guangcheng@ucla.edu}
}

\begin{document} 

\maketitle

\begin{abstract}
Devising procedures for downstream task-oriented generative model selections is an unresolved problem of practical importance. Existing studies focused on the utility of a single family of generative models. They provided limited insights on how synthetic data practitioners select the best family generative models for synthetic training tasks given a specific combination of machine learning model class and performance metric. In this paper, we approach the downstream task-oriented generative model selections problem in the case of training fraud detection models and investigate the best practice given different combinations of model interpretability and model performance constraints. Our investigation supports that, while both Neural Network(NN)-based and Bayesian Network(BN)-based generative models are both good to complete synthetic training task under loose model interpretability constrain, the BN-based generative models is better than NN-based when synthetic training fraud detection model under strict model interpretability constrain. Our results provides practical guidance for machine learning practitioner who is interested in replacing their training dataset from real to synthetic, and shed lights on more general downstream task-oriented generative model selection problems.
\end{abstract}


\section{Introduction}
\label{sec:intro}

Synthetic Data is receiving increasing attention from both academics and industry \cite{jordon2022synthetic}.
Such attention is due to synthetic data's potential to accelerate innovation and support decision-making without violating modern privacy regulations (e.g. GDPR \cite{EUdataregulations2018} or CCPA). The potential for acceleration and support of the modern machine learning lifecycle is lucrative and hence invites both machine learning researchers and practitioners investigation on the benefits and limitations of using synthetic datasets \cite{visani2022enabling, el2020practical}. In particular, 
\textbf{what is the prize and price on \textit{machine learning model training process} if we replace from real training dataset with a synthetic training dataset?}

Indeed, the prize is individual-level privacy protection, and the price is performance degradation. Although the current community does not reach an agreement on how effective the synthetic data approach \cite{stadler2022synthetic}, existing studies \cite{hittmeir2019utility} observe performance degradations from replacing real with the synthetic training datasets. 
In practice, there are three family of generative models to synthesize training dataset: Machine Learning-based \cite{eno2008generating, caiola2010random, drechsler2010using}, Neural network-based \cite{park2018data, xu2018synthesizing} and Bayesian network-based \cite{zhang2017privbayes, gogoshin2021synthetic, abowd2008protective} generative models. 
Due to the predicament of striking a balance between accuracy and privacy leakage for Machine Learning-based generative models \cite{jordon2022synthetic}, in this paper, we focus on Neural network-based and Bayesian network-based generative models.
Given primitive privacy protection by replacing the real training dataset with the synthetic dataset, the following question guides our investigation:
\textbf{Which family of synthetic data generative models sufferers least performance degradation?}
We define the above key research question as the \textbf{Generative Model Selection (GMS)} problems.
In particular, we investigate the generative model selection problem in the context of the fraud detection model (FDM) training procedure.

% ...
The fraud detection model is an integral part of the modern fraud management process \cite{wilhelm2004fraud, abdallah2016fraud,ryman2018artificial}, where the utility-interpretability trade-off is of particular importance to stop fraudsters to limit fraud impact in financial service operations \cite{nesvijevskaia2021accuracy}.
The utility is the key performance metric to measure how effective the trained fraud detection model detecting potential fraudulent operations. However, FDM with high utility also incurs a high number of false alerts, leading to potential high human resource cost\cite{nesvijevskaia2021accuracy}. In practice, those alerts reported by FDM are reviewed by fraud experts, demanding model interpretability to make decisions on free or suspend the potential fraudulent operation. Consequently, \textbf{insights on the impact of synthetic dataset trained FDM, especially around the utility-interpretability trade-off, are desired and of practical importance}. 
In answering GMS problems in synthetic dataset trained FDM, this paper aims to  solve the following two questions:
\begin{itemize}
\item \textbf{Utility-oriented GMS}: \textit{Given a cost-specific metric}, which family of generative models suffers the least degradation? 
\item \textbf{Interpretability-oriented GMS}: \textit{Given a specific interpretability constraint}, which family of generative models suffers the least degradation? 
\end{itemize}

\subsection{Contributions}
In this paper, we adopt a downstream task-oriented approach to the evaluation of generative models introduced in section \ref{subsec:SyntDataGeneModel}. Instead of assessing the generative data distribution by looking into the correlation or distance between real and synthetic datasets, we examine
the performance of fraud detection models trained on Neural network-based or Bayesian Network-based synthetic training data. Our evaluation is from 3 dimensions: data (Synthetic Data generative models), models (fraud detection models), and metrics (accuracy, AUROC, recall, precision). Our examination focus on two class of generative model selection problems: Utility-oriented generative model selections and Interpretability-oriented generative model selections.

We provide 3 guidance for fraud data scientists and machine learning practitioners interested in the utility-interpretability behind using synthetic training data to train their fraud detection models 
\begin{itemize}
\item We systematically compare different performance metrics for fraud detection models to give the best advice on different priorities to stop fraudsters.
\item We systematically compare different machine learning model candidates for fraud detection models from the layer of interpretability. 
\item We give guidance to select synthetic data generative models for different model-metric combinations.
\end{itemize}

\textbf{Utility-oriented Generative Model Selections}
The key question of the Utility-oriented generative model selection problem is on which generative model should be used to generate synthetic training data to train fraud detection under \textit{given metric}. We found the best choice of the generative model family is \textit{metric-dependent}. 
In short, we provide \textit{insights on Utility-oriented GMS.(Section \ref{subsec:UOGMS})}


\begin{itemize}
\item Accuracy  did not show a preference between neural network-based (NN-based) and Bayesian network-based (BN-based) models. 
\item AUROC and Recall prefer NN-based generative models.
\item F1 score and Precision prefers BN-based generative models.
\end{itemize}

\textbf{Interpretability-oriented Generative Model Selections}
On the other hand, the interpretability-oriented generative model selection problem asks for the best family of generative models to generate training data \textit{given 
fraud detection model class}. We found that the Bayesian Network-based method is better for an intrinsic interpretable model class, while both generative model families are good for the complex model classes. In short, we provide 
\textit{insights on Interpretability-oriented GMS. (Section \ref{subsec:IOGMS}).}


\begin{itemize}
\item  Intrinsic interpretable model class and medium  interpretable model class prefers BN-based generative models.
\item  Not-Easy interpretable model class shows no preference between NN-based and BN-based models.
\end{itemize}


\subsection{Paper Organization}

This paper is structured as follows.
Section \ref{sec:relatedWork} gives a comprehensive reviews on related research communities across synthetic data generative models (Section \ref{subsec:SyntDataGeneModel}), fraud detection model interpretability (Section \ref{subsec:FDinterpretability}) and fraud detection metric utility (Section \ref{subsec:FDutility}). Section \ref{sec:ExpSetup} gives experiment details on how to synthesize training dataset (Section \ref{subsec:TrainDataSynthesis}), choices of fraud detection model class (Section \ref{subsec:ChoiceFDMC}) and different fraud detection utility metrics (Section \ref{subsec:ChoiceFDutilityMetric}). Section \ref{sec:Evaluation} reports result of our evaluation on how generative models has effect on the imbalance of real training dataset (Section \ref{subsec:CompOrgSyn}), and results on Utility-oriented GMS (Section \ref{subsec:UOGMS}) and Interpretability-oriented GMS (Section \ref{subsec:IOGMS}). Section \ref{subsec:Conclu} talks about our conclusion, new concerns and future direction.

\section{Relate Work}
\label{sec:relatedWork}

\subsection{Synthetic Data Generative Models}
\label{subsec:SyntDataGeneModel}

\textbf{Neural Network-based generative models.}
\textit{Synthetic Data Vault} (SDV) \cite{patki2016synthetic} provides three neural network-based generative models to synthesize data from a single table. Generative adversarial networks (GANs) are commonly used tools for fraud detection synthetic data due to their ability to address imbalanced datasets via data augmentation \cite{charitou2021synthetic, langevin2021synthetic, langevin2022generative}. SDV provides a GAN-based generative model proposed in \cite{xu2019modeling}, namely conditional tabular GAN (CTGAN). CopulaGAN, another GAN-based model included in SDV, is a variation of the CTGAN Model which takes advantage of the cumulative distribution function (CDF) based transformation. Another type of neural network-based generative model is based on variational autoencoders (VAEs) \cite{kingma2013auto}. SDV also provides a VAE-based generative model, namely, TVAE \cite{xu2019modeling}, to synthesize tabular data. Besides neural network-based generative models, SDV offers GaussianCopula  \cite{masarotto2012gaussian} to model the covariances between features in addition to the distributions \cite{llugiqi2022empirical}.

\textbf{Bayesian Network-based generative models.}
\textit{DataSynthesizer} (DS) \cite{ping2017datasynthesizer} has three modes to invoke modules: random mode, independent attribute mode, and correlated attribute mode. The correlated attribute mode uses the GreedyBayes algorithm to construct Bayesian networks to model correlated attributes, which helps to retain the correlation among variables. Another important parameter in DS is epsilon  which represents differential privacy to address data protection and privacy issues. When epsilon approaches 0, the presence or absence of a single case in the input will be undetectable in the output.


\subsection{Fraud Detection Model Interpretability}
\label{subsec:FDinterpretability}


\textbf{(i) Intrinsic Interpretable Model class.}
Logistic Regression \cite{berkson1944application} provides predictions based on the estimated probability of an event occurring. A transaction will be predicted as a fraud if its estimated probability of being a fraud passes some threshold. Decision Tree \cite{breiman2017classification} is one of the non-parametric supervised learning models used for classification tasks. It predicts classes of transactions via decision rules for data features. 
K-nearest Neighbors (KNN) \cite{altman1992introduction, fix1989discriminatory} is another commonly used non-parametric supervised learning method. Under the assumption that similar points can be found near each other, it uses proximity to make predictions about the class of a transaction.

\textbf{(ii) Medium Interpretable Model class.}
Na\"ive Bayes \cite{zhang2004optimality} is a probabilistic classifier based on Bayes' Theorem with the assumption of conditional independence between each pair of data features. The predicted class of a transaction is with the maximum probability. Support Vector Machine (SVM) \cite{Cortes2004SupportvectorN} is designed to find the hyperplane in an n-dimensional space (n is the number of features) that distinctly classifies data points. Random forest \cite{ho1995random} is one of the ensemble learning methods. It is composed of abundant decision trees and aggregates all predicted classes of decision trees to identify the most popular result as the prediction.

\textbf{(iii) Not-Easy Interpretable Model class.}
The generalized Additive Model (GAM) \cite{hastie1987generalized} is a generalized linear model in which the response is linearly dependent on smooth functions. The smooth relationships between the response and each feature can be estimated simultaneously, and the response in test data can be predicted by adding them up. For binary classification, a logit link is applied to data fitting. Xgboost \cite{chen2016xgboost} is a scalable, distributed gradient boosting system under the Gradient Boosting framework. It provides parallel tree boosting and improves computational efficiency and model performance. Neural Additive Model (NAM) \cite{agarwal2020neural} focuses on a linear combination of deep neural networks that each has a single input feature. The model is fitted via training jointly these neural networks and learning complicated relationships between their inputs and outputs.

\subsection{Fraud Detection Metric Utility}
\label{subsec:FDutility}

\textbf{Fraud Management Process.} Fraud management process \cite{wilhelm2004fraud} has achieved great success in financial service industries due to its capability to catch fraudulent transactions \cite{nesvijevskaia2021accuracy}. As the fraud detection models flag some transaction to be suspicious, it may cause interpretability and the fraud agent have difficulty telling whether the flag is a false alert or not. The fraud detection interpretability issues exist in all kinds of financial service applications including credit or debit cards, payments, and loan approval. 


One major reason behind this communication bottleneck in the fraud management process is the interpretability of detection. As the fraud data scientist wishes to catch as more suspicious operations as possible, the fraud detection model itself becomes more difficult to interpret by using a more complex model class, e.g. neural network-based model. Consequently, the alerted operations become difficult to interpret, and the fraud experts need to take more time to identify the fraud factor, leading to such a communication bottleneck. Such a late decision allows the fraudsters to maximize their fraud fain, which greatly impairs the effectiveness of the model and also results in enormous economical and opportunity loss. As such, the interpretability in fraud detection model machine learning has become a rising concern.

In order to trade off the model performance and model interpretability, existing works are all focused on a model-centric approach. The model-centric approach takes the mindset of "accuracy first, interpretability later". They first compare the performance of different classes of fraud detection models. Then adopts the model's nature to gain the model interpretability. In addition to the intrinsic interpretable model class, there are various methods proposed to do post-doc methods to gain model-specific interpretability. The main drawback of the model-centric approach is the lack of a fair way to compare model interpretability in the same framework.

\textbf{Performance Metrics.} We evaluate synthetic data training of FDM with metrics for standard classifier and for imbalanced dataset trained classifier. 
\textbf{(i) Metrics for standard classifier.}
Accuracy \cite{powers2020evaluation} is one of the commonly used metrics in machine learning tasks. It is useful when all classes are of equal importance, but it can be misleading for an imbalanced dataset. AUROC \cite{hanley1982meaning} describes the model’s ability to discriminate between positive cases and negative cases. F1 score \cite{sasaki2007truth} measures the performance of a model by computing the harmonic mean of the precision and recall of the model. \textbf{(ii) Metrics for imbalanced dataset trained classifier.}
\textit{Recall} and \textit{precision} \cite{powers2020evaluation, allen1955machine} are two metrics which are more suitable than accuracy for imbalanced testing datasets. There is an inverse relationship between precision and recall usually. Precision-Recall curve \cite{raghavan1989critical} shows the tradeoff between precision and recall for different thresholds. Average precision \cite{schutze2008introduction} summarizes such a tradeoff as the weighted mean of precisions achieved at each threshold, where the weight is the increase in recall from the prior threshold.

\section{Experiment Setup}
\label{sec:ExpSetup}

\subsection{Training Data Synthesis}
\label{subsec:TrainDataSynthesis}

\textbf{Dataset-Credit Card Fraud Dataset.} We conduct the experiment on the Credit Card Fraud Dataset \cite{dal2015calibrating} which is highly imbalanced (0.1727\% transactions are frauds). The dataset contains 31 variables. The feature "Amount" is the transaction amount, and the feature "Time" represents the seconds elapsed between each transaction and the first transaction in the dataset. "Class" is the response variable and it takes 1 for fraudulent cases and 0 for other cases. Due to confidentiality, the remaining 28 features are principal components obtained by the means of PCA. In our experiment, we delete the feature "Time" since it has nothing to do with "Class".



\textbf{Generating Synthetic Training Dataset.}
SDV is applied with GaussianCopula, CTGAN, CopulaGAN and TVAE, to generate synthetic data. We employ the four generative models on the original dataset respectively and obtain 4 synthetic datasets. DS is utilized with the correlated attribute mode. Synthetic datasets are generated with two generative models. One is with epsilon = 0, and the other is with epsilon = 0.1.

Since the original dataset is highly imbalanced, some synthetic data generative models cannot yield transactions with class 1 (fraudulent cases). It is found that no fraudulent transactions are synthesized when using GaussianCopula and CopulaGAN. Therefore, we leave out the datasets generated by GaussianCopula and CopulaGAN in the following tasks. 



\subsection{Choice of Fraud Detection Model Class}
\label{subsec:ChoiceFDMC}

\textbf{Synthetic Data Training Procedure.}
After generating synthetic datasets of the same size as the original dataset, we randomly split the original dataset into training data (70\% of the original data) and test data (30\% of the original data), and randomly select 70\% of each synthesized dataset as the synthesized training data. Then we fit fraud detection models to the training data and each synthesized training data and predict the results on the test data. 

\textbf{Model class by Interpretability.} We fit 9 fraud detection models to analyze utility-interpretability on FDM training tasks. The following list summarizes 9 models class from high to low interpretability:
\begin{itemize}
\item \textbf{Intrinsic interpretable models}: Logistic Regression (LR),  Decision Tree (DT) and K-nearest Neighbors (KNN).
\item \textbf{Medium interpretable models}: Na\"ive Bayes (NB), Support Vector Machines (SVM), Random Forest (RF).
\item \textbf{Not-easy interpretable models}: Generalized Additive Model (GAM),  Extreme Gradient Boosting (XGBoost), Neural Additive Model (NAM).
\end{itemize}


\subsection{Choice of Fraud Detection Utility Metrics}
\label{subsec:ChoiceFDutilityMetric}

To evaluate the performance of synthesized data on fraud detection, we compute accuracy, AUROC, recall, precision, and F1 score, and generate the Precision-Recall curve and average precision (AP) for each machine learning task based on the predicted results.

\textbf{Basic Performance Metrics for Classifier.}

\begin{itemize}
    \item \textbf{Accuracy} is the ratio of the number of correct predictions to the number of all predictions. It describes how the model performs across both classes.
    \item \textbf{AUROC} measures the ability of a classifier to distinguish between the fraudulent class and the nonfraudulent class.
    \item \textbf{F1 score} is the harmonic mean between precision and recall. It provides equal importance to precision and recall.
\end{itemize}

\textbf{Performance Metrics for Classifier on Imbalanced Dataset.}

\begin{itemize}
    \item \textbf{Precision} is defined as the ratio of true positives to the sum of true positives and false positives. It is the number of correctly predicted fraudulent cases divided by the total number of predicted fraudulent cases for fraud detection.
    \item \textbf{Recall} is defined as the ratio of true positives to the sum of true positives and false negatives. It is the number of correctly predicted fraudulent cases divided by the total number of actual fraudulent cases in the fraud detection scenario.
    \item \textbf{Precision-Recall curve} shows the tradeoff between precision and recall for different thresholds. A high area under the curve represents both high recall and high precision.
    \item \textbf{Average precision} is calculated as the weighted mean of precisions at each threshold, where the weight is the increase in recall from the prior threshold.
\end{itemize}


\section{Evaluation}
\label{sec:Evaluation}

\subsection{Comparison of Original to Synthetic Data}
\label{subsec:CompOrgSyn}
\textbf{(1) Balance.}
Table \ref{tab:synthetic_balance} shows the degree of imbalance of synthetic training datasets.
Datasets generated by CTGAN and DS with epsilon = 0.1 are balanced (59.4919\% and 45.4743\% transactions are frauds respectively). TVAE synthesizes a more imbalanced dataset (0.0119\% transactions are frauds) than the original dataset. When using DS with epsilon = 0, the synthesized dataset contains 0.1731\% fraudulent transactions, which is similar to the original dataset.

\begin{table}[]
\begin{center}
    \caption{Percentage of classes}
    \label{tab:synthetic_balance}
    \begin{tabular}{ccl}
    \toprule
    Approach & Class 1 (frauds) & Class 0 \\
    \midrule
    Original & 0.1727\% & 99.8273\% \\
    CTGAN & 59.4919\% & 40.5081\%\\
    TVAE & 0.0119\% & 99.9881\% \\
    DS 0 & 0.1731\% & 99.8269\%\\
    DS 0.1 & 45.4743\% & 54.5257\% \\
    \bottomrule
    \end{tabular}
\end{center}
\end{table}

\textbf{(2) Correlation.}
 Dataset generated by DS with epsilon = 0 maintains the correlation of each pair of two features. However, datasets synthesized by other generative models do not retain the correlation between features.




\subsection{Results on Utility-oriented GMS}
\label{subsec:UOGMS}



In order to compare the performance of the original data and synthesized data in fraud detection tasks, we generate line charts to show the results of each metric, for data generated with different approaches as models' interpretability increases.

\begin{figure*}[t]
    \includegraphics[width=.32\textwidth]{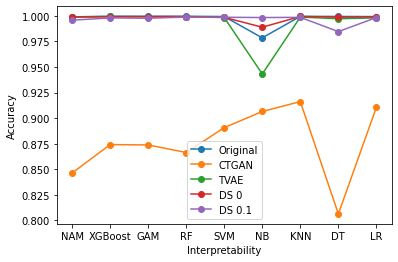}
    \includegraphics[width=.32\textwidth]{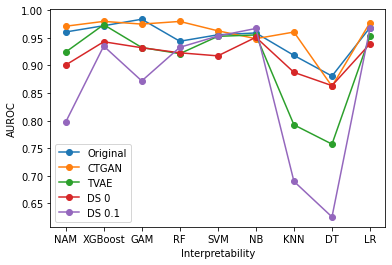}
    \includegraphics[width=.32\textwidth]{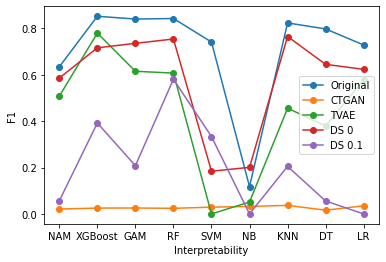}
    \includegraphics[width=.32\textwidth]{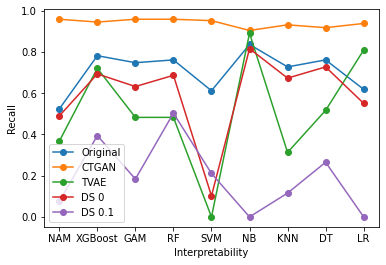}
    \includegraphics[width=.32\textwidth]{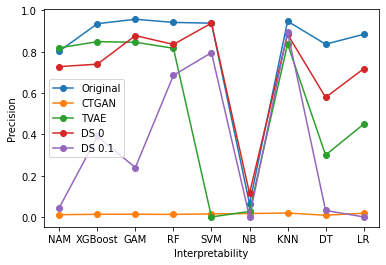}
    \caption{Results to solve Utility-oriented GMS: Utility Metrics for Fraud Detection Classifiers}
    \label{metrics}
\end{figure*}

\textbf{(1) Accuracy.}
Since the original dataset is highly imbalanced, the accuracy on test data is very close to 1 for each fraud detection model. In Figure \ref{metrics}, \textit{training dataset generated by TVAE, DS with epsilon 0 and DS with epsilon 0.1 also yield high accuracy}. The accuracy of data synthesized by DS is even higher than the accuracy of the original data for Na\"ive Bayes. CTGAN has lower accuracy than other approaches for all fraud detection models. 

It is remarkable that
\textbf{accuracy of Na\"ive Bayes is highly unstable across different synthetic training datasets}, while other fraud detection models are with stable accuracy except for CTGAN-based training data. Although DS with epsilon 0.1 has the highest accuracy for Na\"ive Bayes, its accuracy for Decision Tree is the second lowest. Compared with the original data, data synthesized by DS with epsilon 0 has almost the same or higher accuracy. 

Hence, \textit{DS with epsilon 0 performs the best in accuracy overall for fraud detection tasks}. Bayesian network-based generative models are selected by accuracy metric due to their better performance in Figure \ref{metrics}, where both Bayesian network-based generative models outstrip neural network-based generative models.

\textbf{(2) AUROC.}
For AUROC, the original data performs the best for GAM and Decision Tree. DS with epsilon 0.1 improves the performance for Na\"ive Bayes, CTGAN and TVAE improve the results for XGBoost, and only CTGAN increases the value of AUROC for the other 5 fraud detection models. DS with epsilon 0 produces smaller AUROC than the original data for all models. \textbf{Na\"ive Bayes and Logistic Regression show robustness while the result of AUROC varies considerably among data generative approaches for KNN and Decision Tree}. Figure \ref{metrics} indicates that CTGAN is a suitable data generative technique to synthesize data considering the performance of AUROC, but for Na\"ive Bayes, DS with epsilon 0.1 is preferred.

In the fraud detection scenario, \textit{Na\"ive Bayes has better discrimination between fraudulent transactions and nonfraudulent transactions when the training data is generated by DS with epsilon 0.1}. When the training data is synthesized by CTGAN, Logistic Regression, KNN, SVM, Random Forest, XGBoost and NAM perform better at distinguishing between fraudulent cases and nonfraudulent cases. Thus, \textbf{AUROC prefers neural-network based generative models, especially CTGAN which achieves comparable performance to the original data}.

\textbf{(3) F1 score.}
According to Figure \ref{metrics}, \textit{only DS with epsilon 0 has higher F1 score than the original for Na\"ive Bayes}. For other fraud detection models, the original data has the best performance, especially for SVM for which there is a noticeable difference in F1 score between the original data and synthesized data. In addition, it is found that \textit{F1 scores of CTGAN are almost 0 for all fraud detection models}. Even though F1 scores are relatively similar for Na\"ive Bayes, none of the models show robustness for all datasets. Data generated by DS with epsilon 0 yields the highest F1 score, but it seems doubtful that this approach is suitable for Na\"ive Bayes because all F1 scores, including the F1 score of DS with epsilon 0, are very low for Na\"ive Bayes.

Combining recall and precision, DS with epsilon 0 improves the performance of Na\"ive Bayes for fraud detection. Other fraud detection models show better performance when they are trained by the original dataset. In conclusion, \textbf{F1 score selects Bayesian network-based generative models since overall DS with epsilon 0 has the best performance among all generative techniques}.

\textbf{(4) Recall.}
\textit{CTGAN has considerably higher recall} than other approaches in Figure \ref{metrics}. Compared with the original dataset, data generated by TVAE also increases recall scores for Na\"ive Bayes and Logistic Regression while DS with epsilon 0 and DS with epsilon 0.1 perform worse than the original for all fraud detection models. \textit{Only Na\"ive Bayes is robust for recall when leaving out DS with epsilon 0.1}. Because of the remarkable performance, \textbf{CTGAN is the suitable technique to generate synthetic data when we focus on recall scores}.

All machine learning models yield more correctly predicted frauds over total actual frauds in the fraud detection tasks when the training data is synthesized by CTGAN. More fraudulent cases are recognized properly when using CTGAN to generate the training data. Therefore, \textbf{neural network-based generative models are preferred by recall} because CTGAN surpasses all of the other methods and even distinctly exceeds the original data.

\textbf{(5) Precision}
Compared with the original, TVAE increases the precision score a little for NAM, and DS with epsilon 0 improves the performance in precision for Na\"ive Bayes in Figure \ref{metrics}. Data synthesized by DS with epsilon 0 produces the same precision score to the original data for SVM. We can see that DS with epsilon 0.1 has worse performance than the original for all fraud detection models, and all precision scores of CTGAN are close to 0. Except for Na\"ive Bayes, all models show differences in precision scores among approaches. For NAM, TVAE is a suitable method for data generation, and for SVM and Na\"ive Bayes, DS with epsilon 0 can synthesize data comparable to the original data when focusing on precision. There are no synthetic data generative models with comparable precision to the original dataset for other machine learning models.

For fraud detection, NAM can correctly predict more frauds among all predicted fraudulent cases when the training data is generated by TVAE. Na\"ive Bayes has more correct predictions for cases predicted to be fraudulent when the training data is synthesized by DS with epsilon 0. Other fraud detection models produce a higher ratio of correct predicted frauds to all fraudulent predictions when it is trained by the original data. Among all data generative tools, DS with epsilon 0 has the best performance overall and CTGAN performs the worst, so \textbf{precision prefers Bayesian network-based generative models}.


\begin{figure*}[t]
    \includegraphics[width=.32\textwidth]{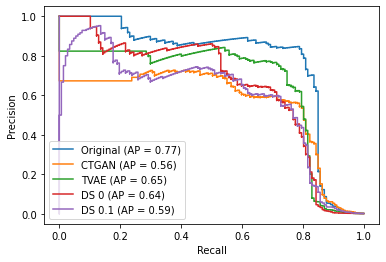}
    \includegraphics[width=.32\textwidth]{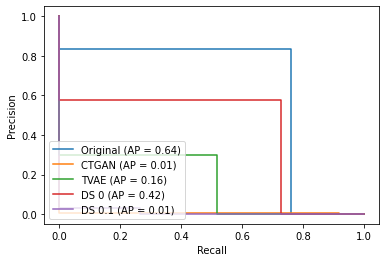}
    \includegraphics[width=.32\textwidth]{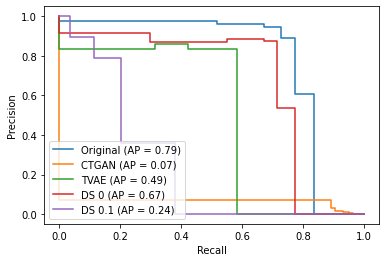}
    \includegraphics[width=.32\textwidth]{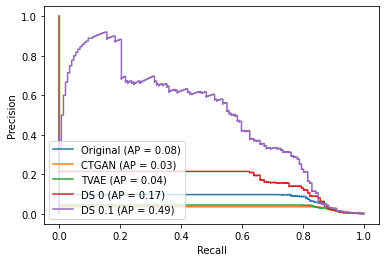}
    \includegraphics[width=.32\textwidth]{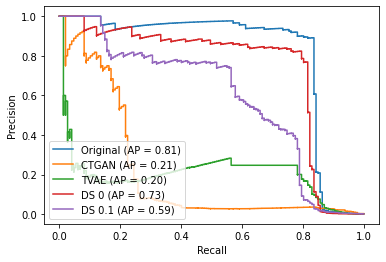}
    \includegraphics[width=.32\textwidth]{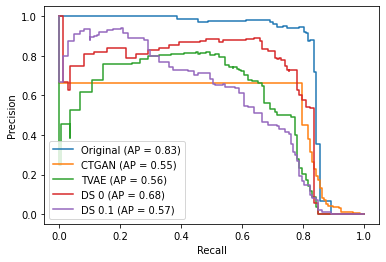}
    \includegraphics[width=.32\textwidth]{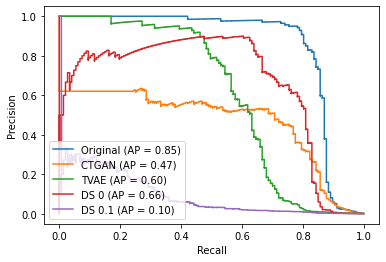}
    \includegraphics[width=.32\textwidth]{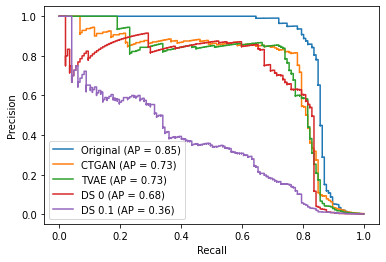}
    \includegraphics[width=.32\textwidth]{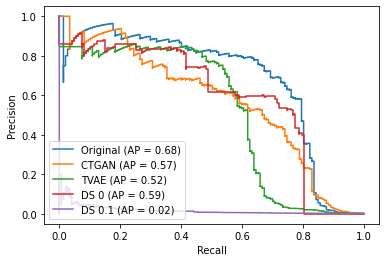}
    \caption{Results to solve Interpretability-oriented GMS: Precision-Recall curve and Average Precision}
    \label{precision-recall}
\end{figure*}

\subsection{Results on 
 Interpretability-oriented GMS}
 \label{subsec:IOGMS}

The original dataset yields Precision-Recall curves in the top right corner and the highest average precision for all machine learning models except for Na\"ive Bayes according to Figure \ref{precision-recall}. DS with epsilon 0.1 has the Precision-Recall curve in the top right corner and the highest average precision for Na\"ive Bayes, followed by DS with epsilon 0. For Logistic Regression, TVAE produces the second highest average precision, and the Precision-Recall curve is just lower than the original. CTGAN and TVAE have Precision-Recall curves just lower than the original and the same average precision for XGBoost. For the other 6 fraud detection models, DS with epsilon 0 yields the second highest average precision, and the Precision-Recall curve is just lower than the original.

\textbf{(i) Intrinsic Interpretable Model class.}
In the first row of Figure \ref{precision-recall}, there is no noticeable difference between neural network-based models and Bayesian network-based models in Precision-Recall curves for Logistic Regression. So, Logistic Regression indicates that neural network-based and Bayesian network-based generative models have comparable performance. Decision Tree and KNN suggest Bayesian network-based generative models as better tools for synthetic data generation. 

\textbf{(ii) Medium Interpretable Model class.}
In the second row of Figure \ref{precision-recall},
\textbf{all medium interpretable models select Bayesian network-based generative models to synthesize data}. Bayesian network-based generative models even have a better performance than the original for Na\"ive Bayes. Although Bayesian network-based generative methods perform better for all medium interpretable machine learning models, the difference is that Na\"ive Bayes prefers DS with epsilon 0.1 while SVM and Random Forest prefer DS with epsilon 0.

\textbf{(iii) Not-Easy Interpretable Model class.}
The third row of Figure \ref{precision-recall} presents the parallel performance of Bayesian network-based and neural network-based generative models for not-easy interpretable fraud detection models. For GAM, TVAE and DS with epsilon 0 show comparable performance. For XGBoost and NAM, CTGAN, TVAE and DS with epsilon 0 have similar Precision-Recall curves. Curves for DS with epsilon 0.1 are located in the bottom left corner for all not-easy interpretable models.

\subsection{Results on Synthetic Augmented Training}
 \label{Sec:SynAuf:Result}

By \textit{synthetic augmented training}, we mean the training Dataset for ML classifier is a mixture of source real dataset and certain percentage of synthetic data. In particular, we consider five differen degree of such real-synthetic mixture: \texttt{syn0.1, syn0.2, syn0.3, syn0.4, syn0.5}. For example, \texttt{syn0.2} means the training dataset is the mixture of 100\% source real dataset and 20\% of synthetic dataset from the synthesizer. 

 The full empirical results on Synthetic Augmented Training from  metric-oriented and synthesizer-oriented are given at section \ref{app:Graph_SynAug}. We summarize key insights below. (1) \texttt{CTGAN}-augmented training datasets improves the AUROC and recall of synthetic trained ML classifier. However, \texttt{CTGAN}-augmented training datasets damages accuracy, F1 score, precision and precision-recall curve across all synthetic trained ML classifier. (2) \texttt{TVAE}-augmented training datasets in general do not improve or damage the utility of synthetic trained ML classifier across all 6 performance performance metrics and all 9 ML Classifier. (3) \texttt{PrivBayes}-augmented training datasets improves the accuracy of synthetic trained ML classifier. However, \texttt{PrivBayes}-augmented training datasets damages AUROC, F1 score, recall, precision across all synthetic trained ML classifier.

\section{Conclusion}
\label{subsec:Conclu}

In this paper, we provide a practical evaluation of generative model selections for synthetic training of fraud detection models. Our evaluation framework covers data, models, and metrics and provides results to answer utility-oriented generative model selections and interpretability-oriented generative model selections. 

One promising future work direction is to develop \textit{generative model auditing process} for generative models and their synthetic datasets. Such model auditing process has been explored in the domain-agnostic and model-agnostic way \cite{alaa2022faithful}, but more task-oriented studies on model generative model auditing process are in demand. Indeed, such a generative model auditing process will generate values for data scientists and machine learning practitioners in integrating synthetic datasets into their daily workflow and leading a more trustworthy machine learning lifecycle in the near future.

\textbf{Disclaimer.} This paper was prepared for informational purposes by the Artificial Intelligence Research group of JPMorgan Chase \& Co and its affiliates (“J.P. Morgan”), and is not a product of the Research Department of J.P. Morgan. J.P. Morgan makes no representation and warranty whatsoever and disclaims all liability, for the completeness, accuracy or reliability of the information contained herein. This document is not intended as investment research or investment advice, or a recommendation, offer or solicitation for the purchase or sale of any security, financial instrument, financial product or service, or to be used in any way for evaluating the merits of participating in any transaction, and shall not constitute a solicitation under any jurisdiction or to any person, if such solicitation under such jurisdiction or to such person would be unlawful.

\clearpage

\baselineskip=13pt
\bibliographystyle{plain}
\nocite{*}
\bibliography{ref}

\clearpage
\appendix

\section{Experiment Results on Synthetic Data Augmented Training}
\label{app:Graph_SynAug}

This section provides additional empirical results for Synthetic Augmented Training discussed at Section \ref{Sec:SynAuf:Result}. 

\begin{figure}[h]
    \includegraphics[width=.50\textwidth]{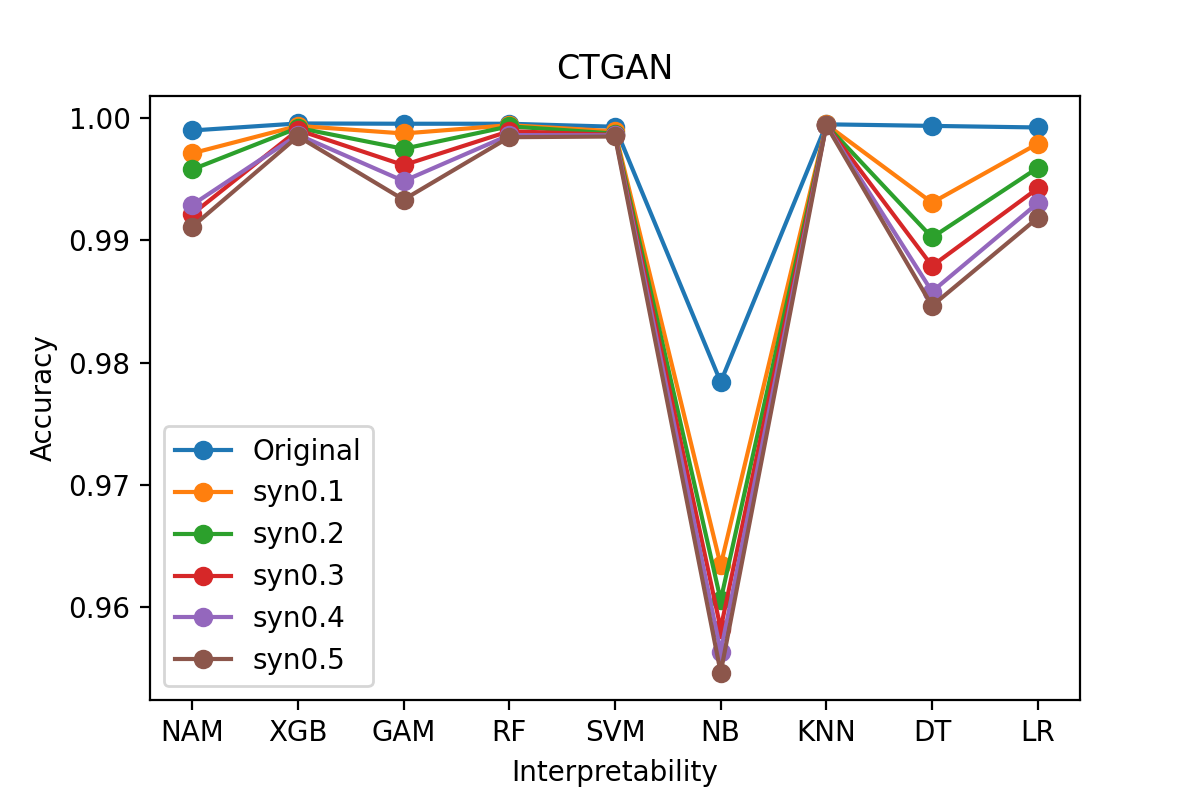}
    \includegraphics[width=.50\textwidth]{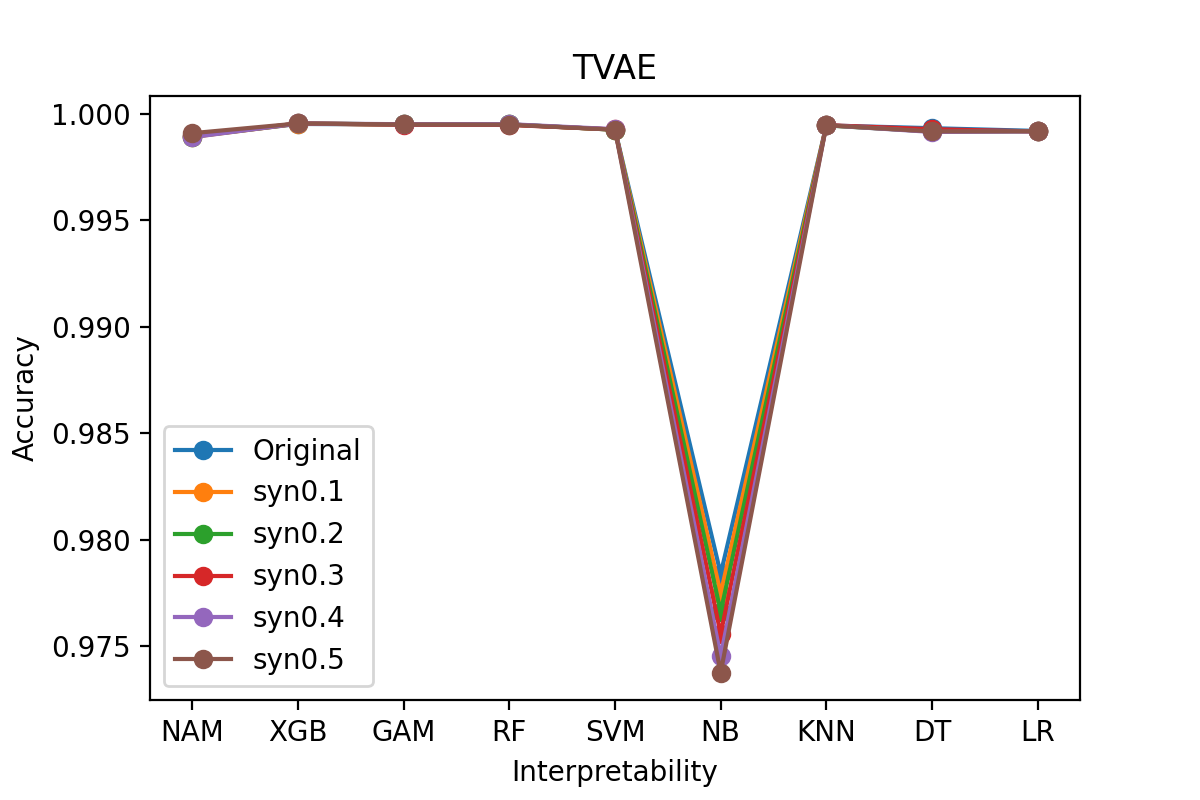}
    \includegraphics[width=.50\textwidth]{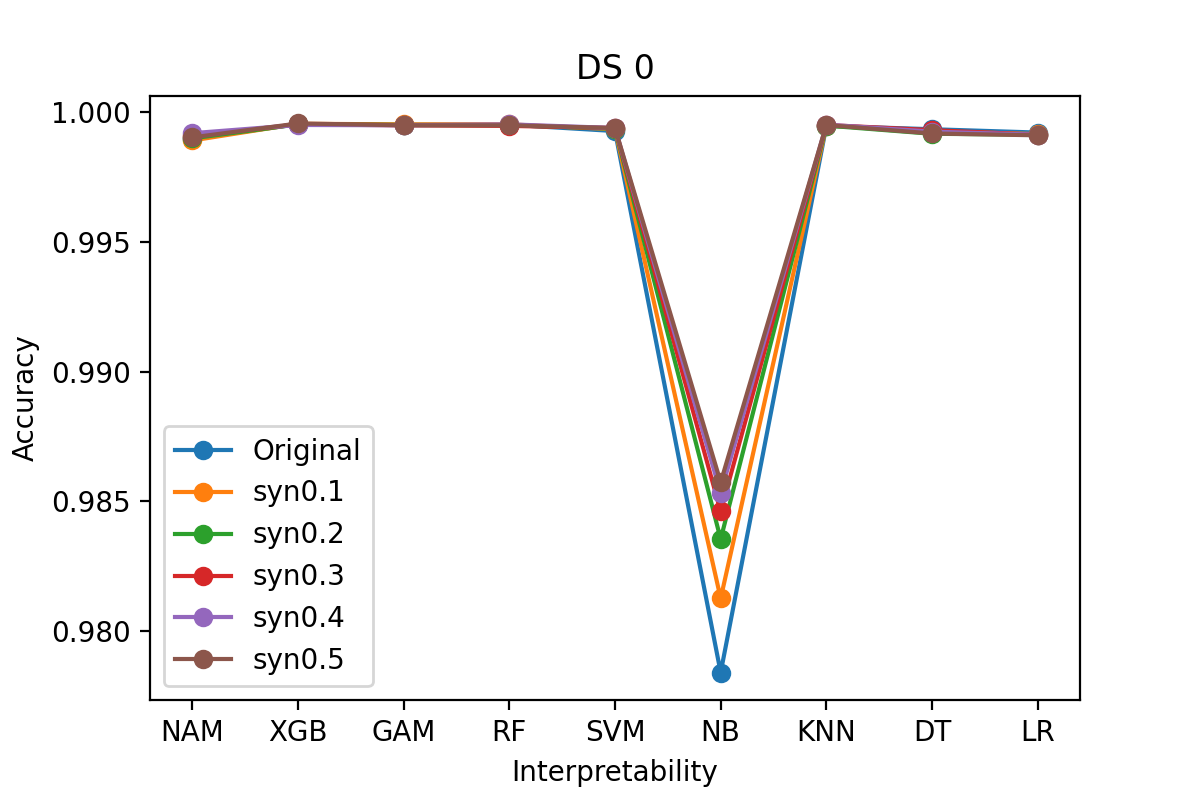}
    \includegraphics[width=.50\textwidth]{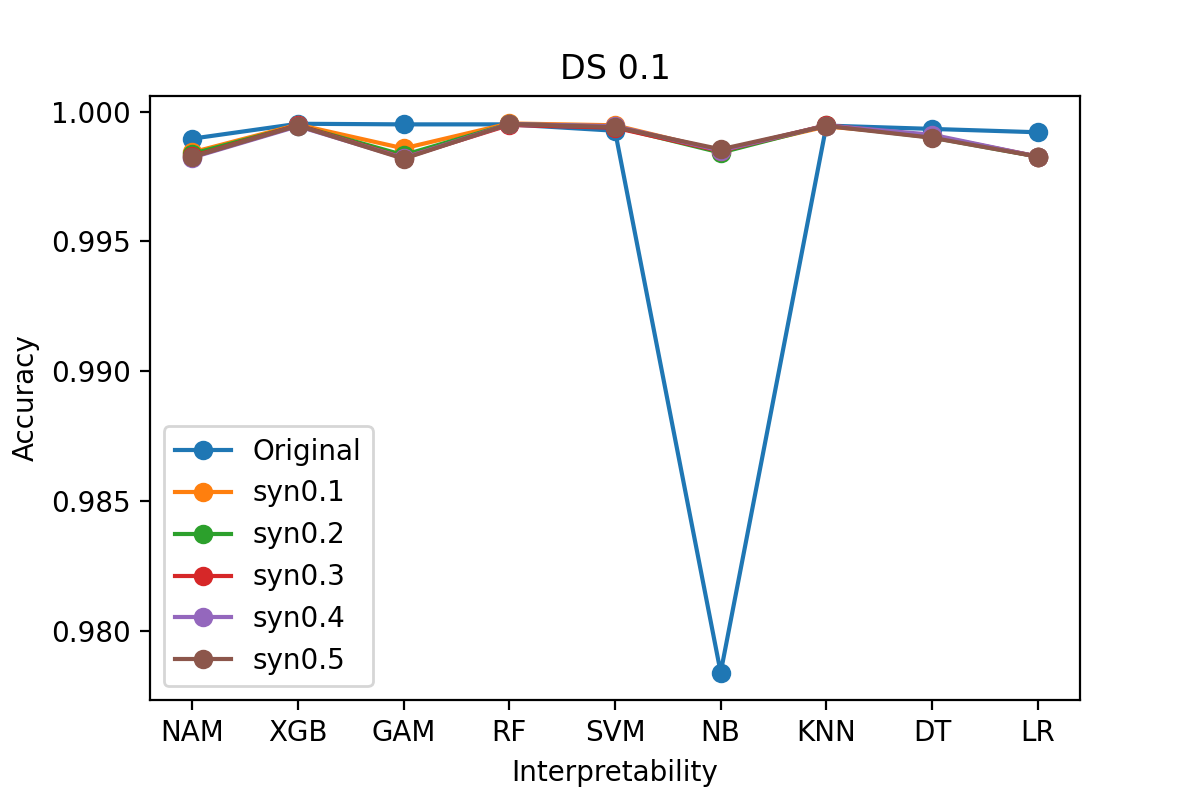}
    \caption{Accuracy. CTGAN-augmented  training dataset damages synthetic trained classifier utility. PrivBayes-augmented training dataset improves synthetic trained classifier utility.}
    \label{SynAug:accuracy}
\end{figure}

\begin{figure}
    \includegraphics[width=.50\textwidth]{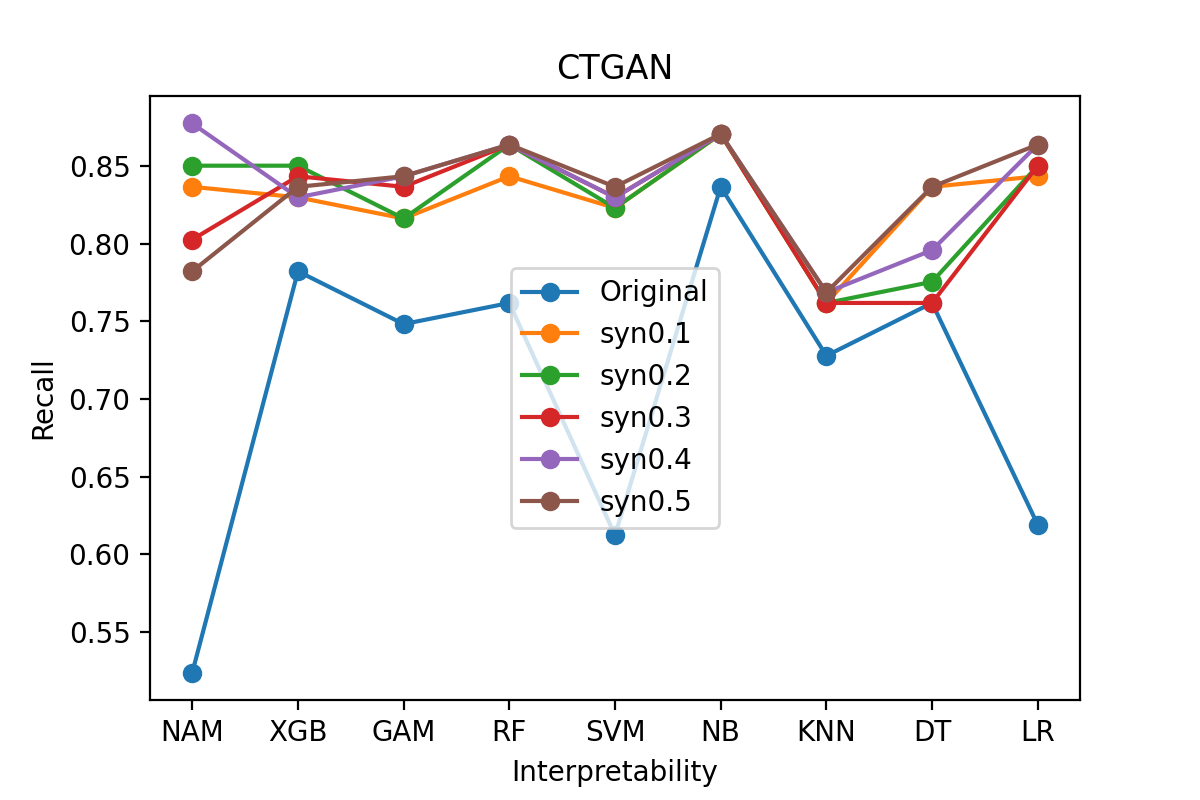}
    \includegraphics[width=.50\textwidth]{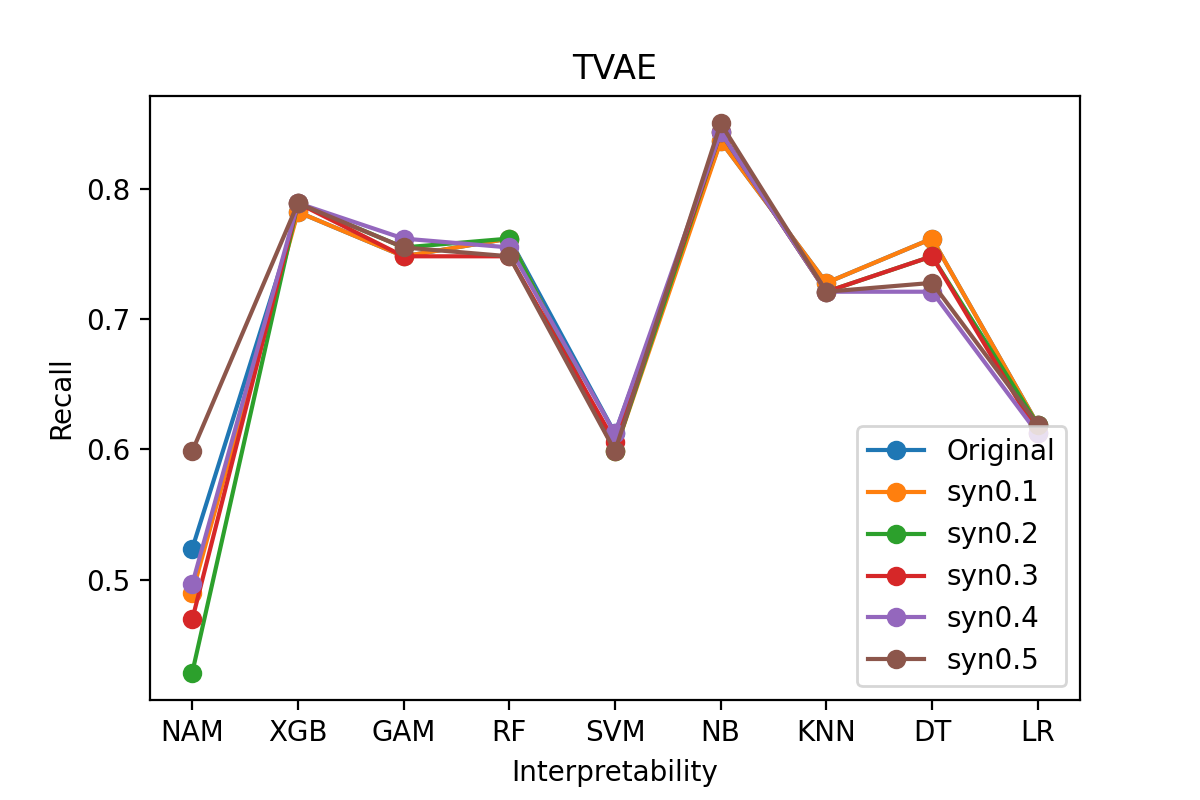}
    \includegraphics[width=.50\textwidth]{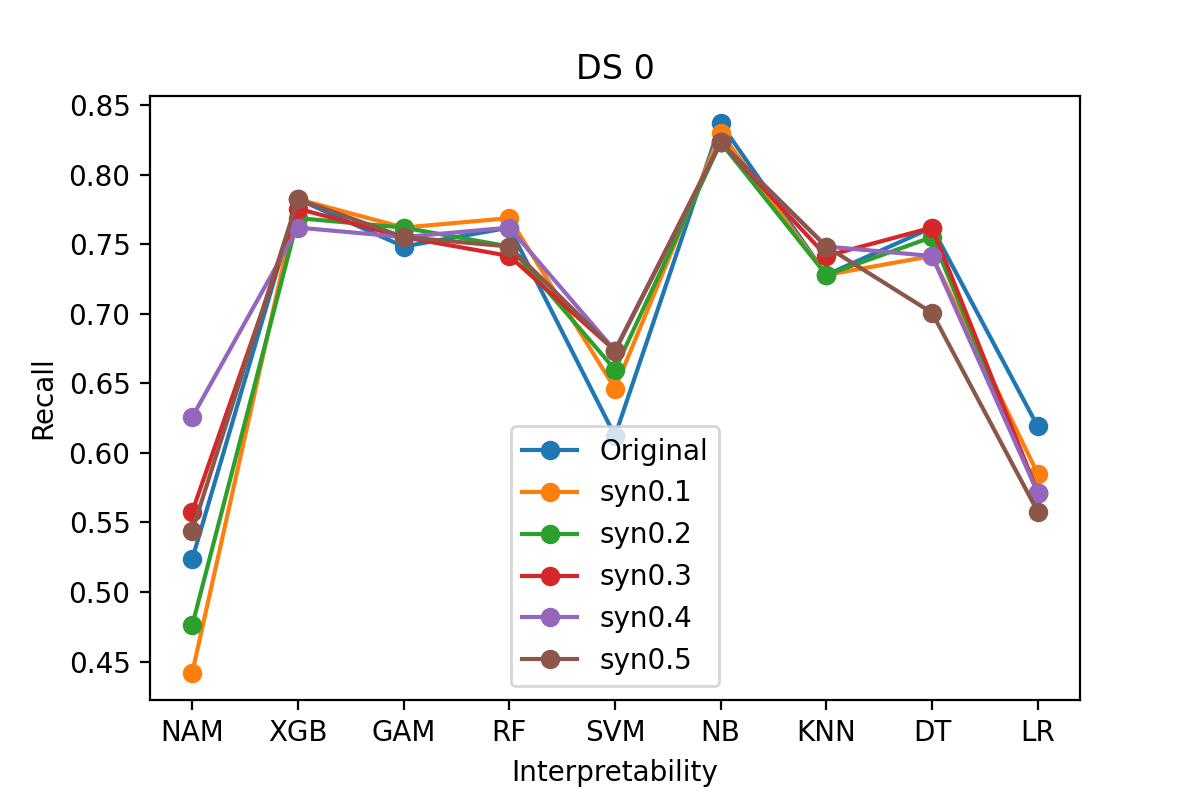}
    \includegraphics[width=.50\textwidth]{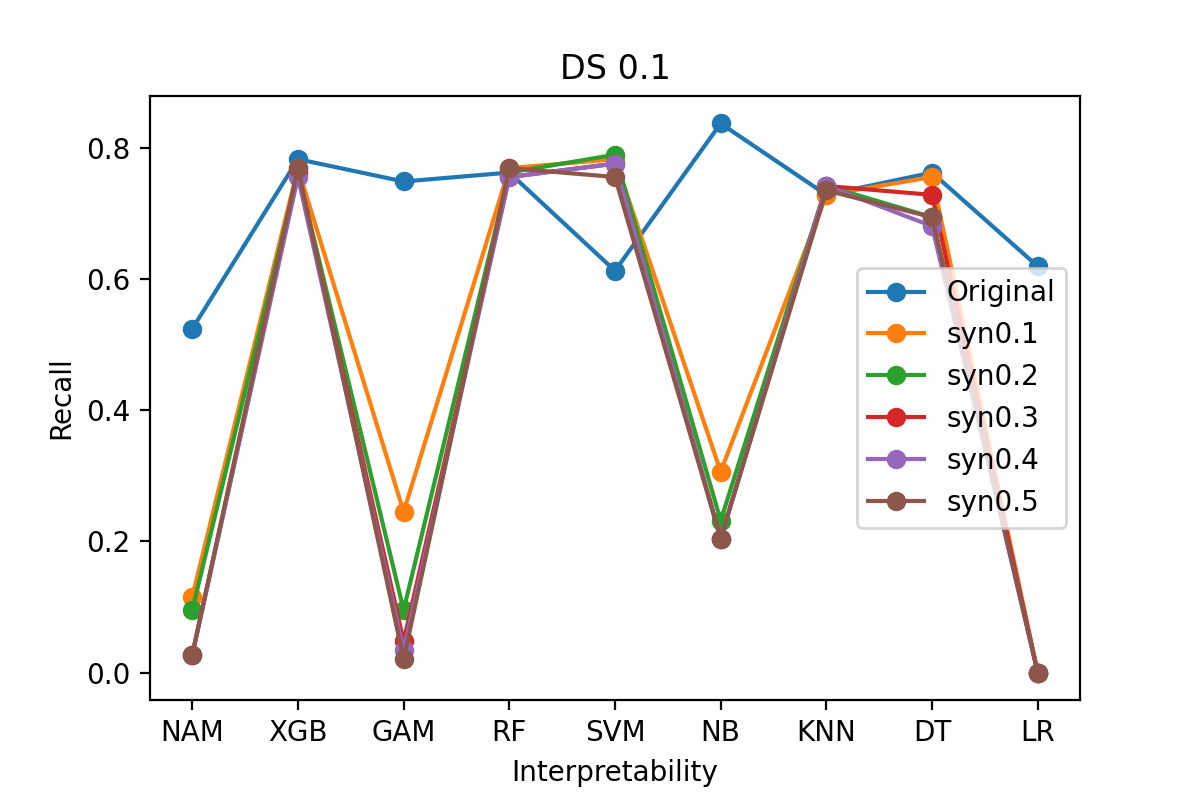}
    \caption{Recall. CTGAN-augmented  training dataset improves synthetic trained classifier utility. PrivBayes-augmented training dataset damages synthetic trained classifier utility.}
    \label{SynAug:recall}
\end{figure}

\begin{figure*}
    \includegraphics[width=.50\textwidth]{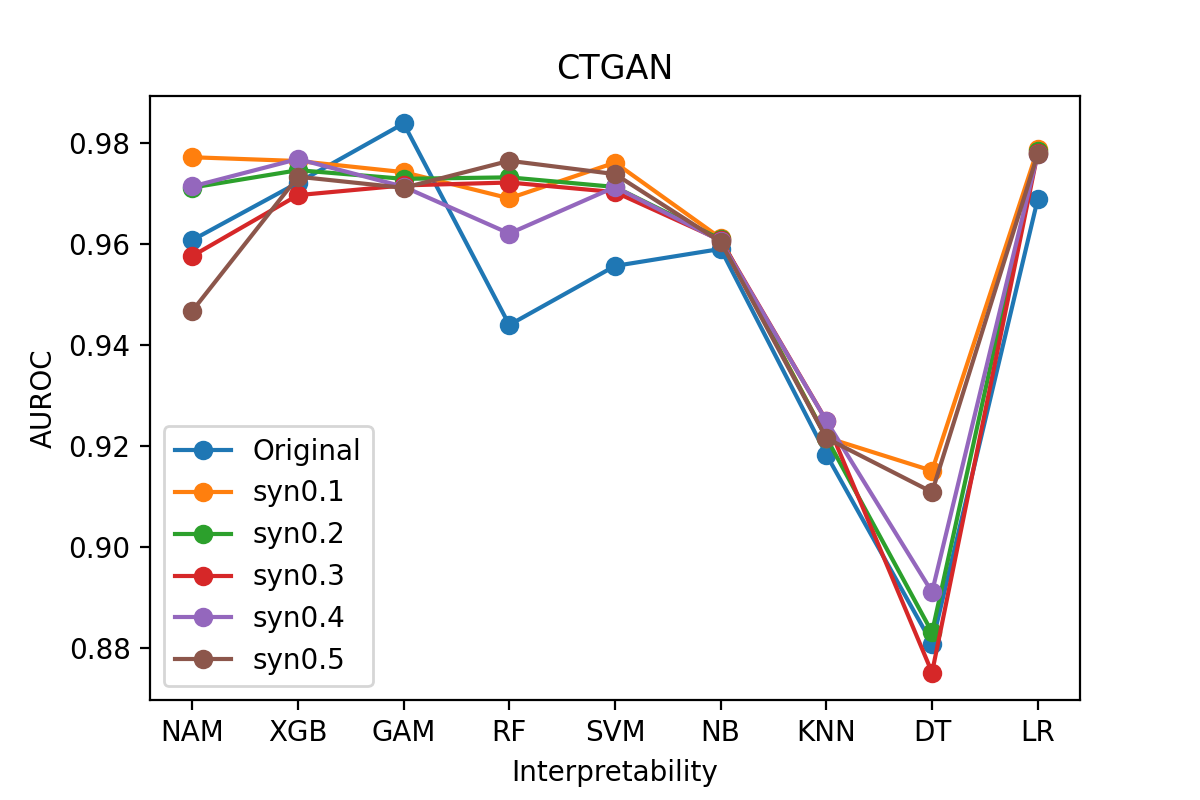}
    \includegraphics[width=.50\textwidth]{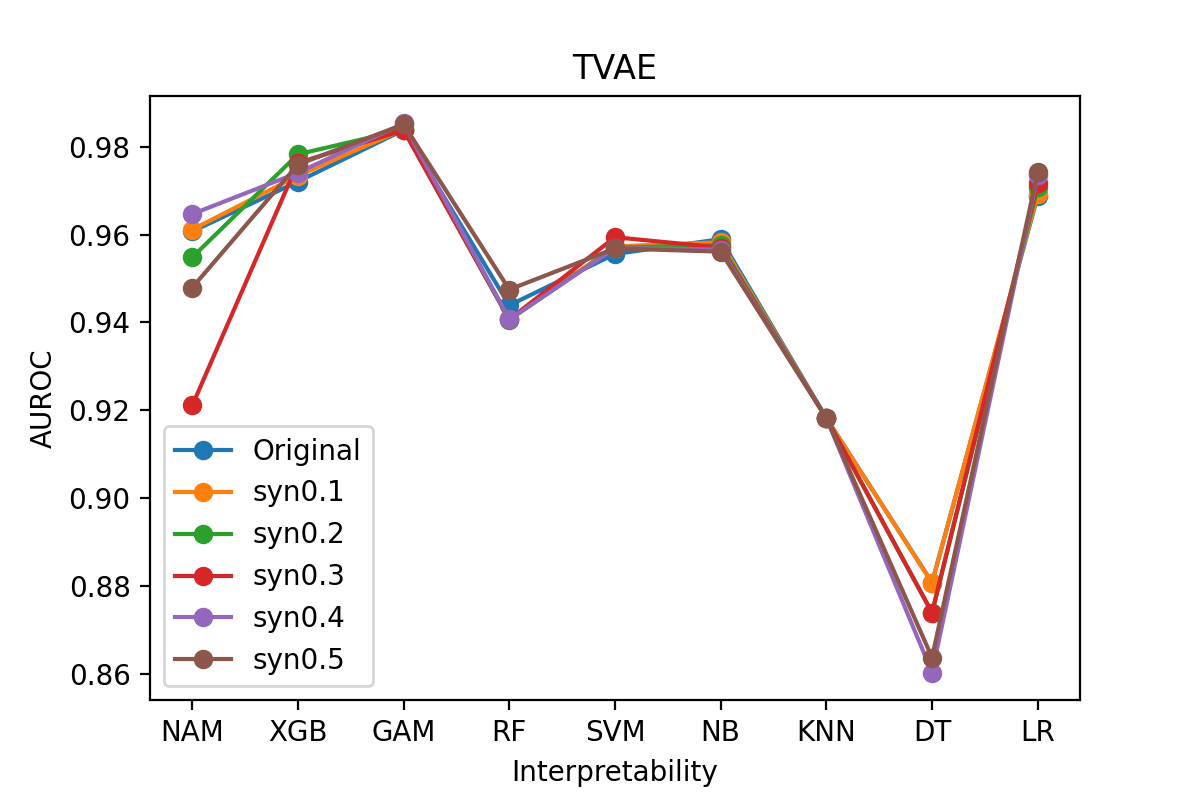}
    \includegraphics[width=.50\textwidth]{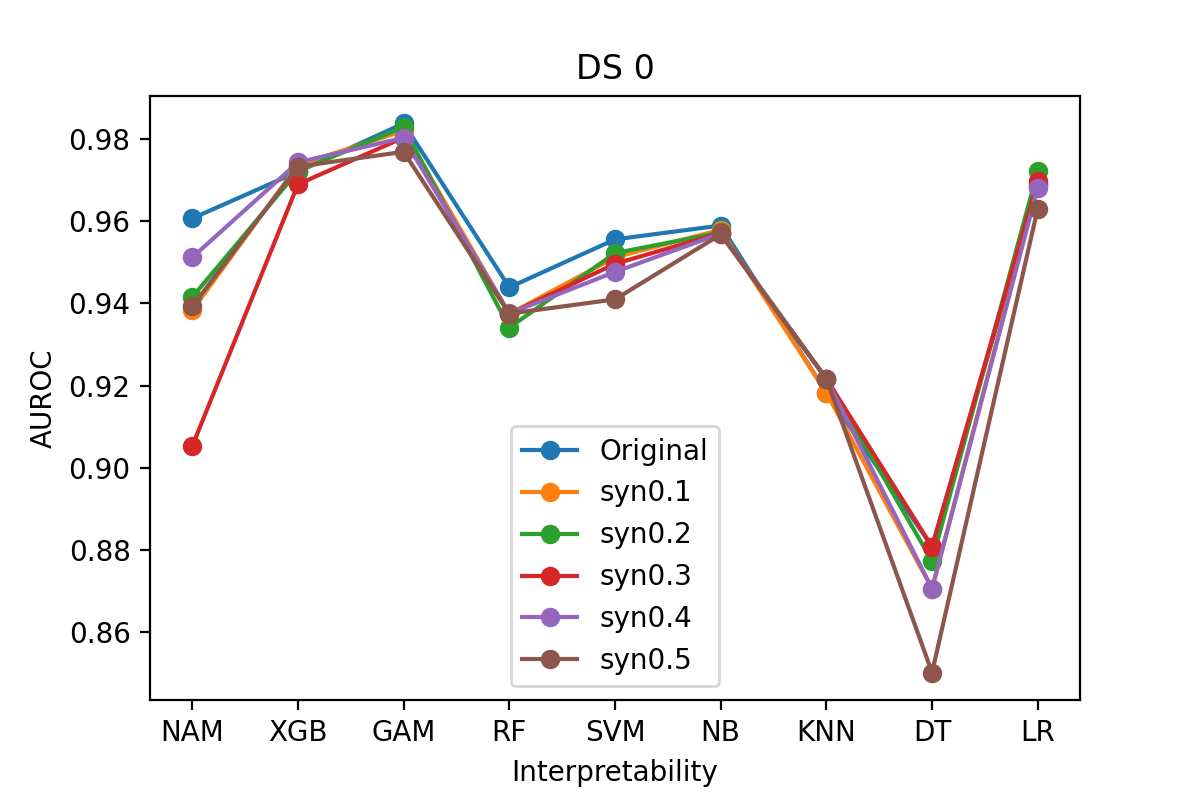}
    \includegraphics[width=.50\textwidth]{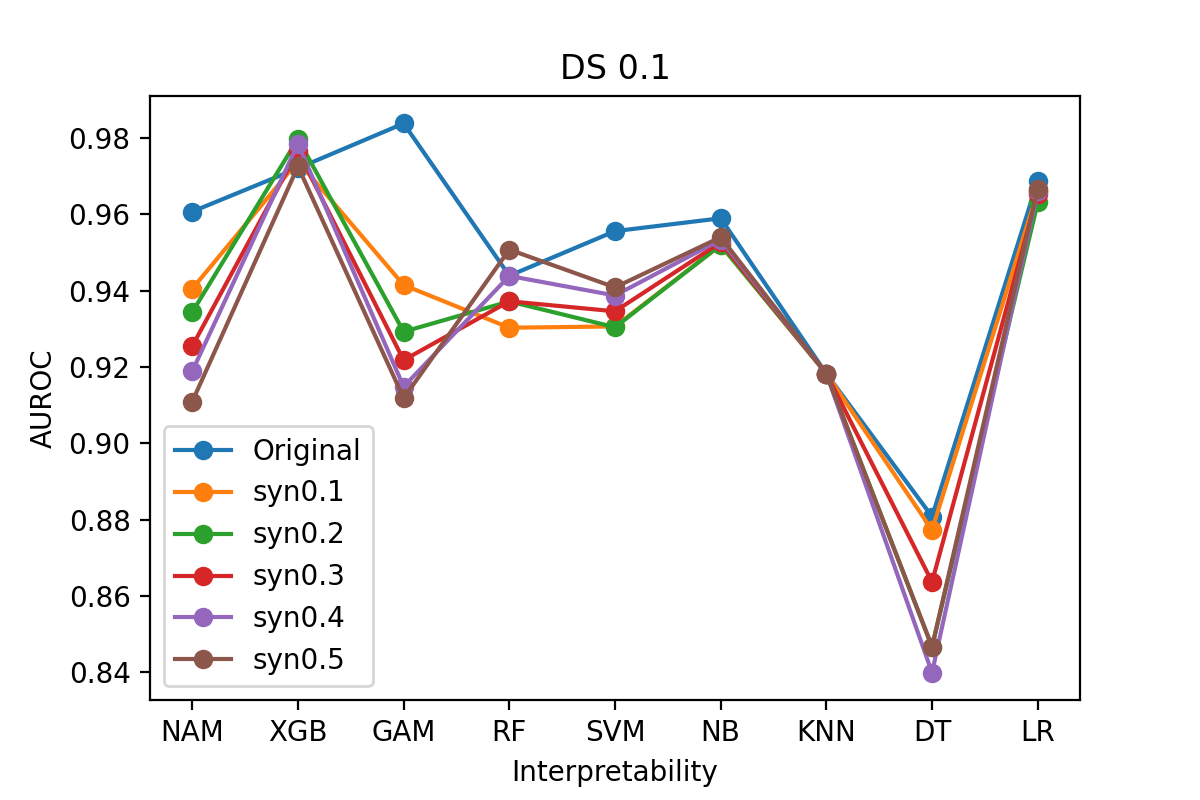}
    \caption{AUROC. CTGAN-augmented  training dataset improves synthetic trained classifier utility. PrivBayes-augmented training dataset damages synthetic trained classifier utility.}
    \label{SynAug:auroc}
\end{figure*}

\begin{figure*}[]
    \includegraphics[width=.50\textwidth]{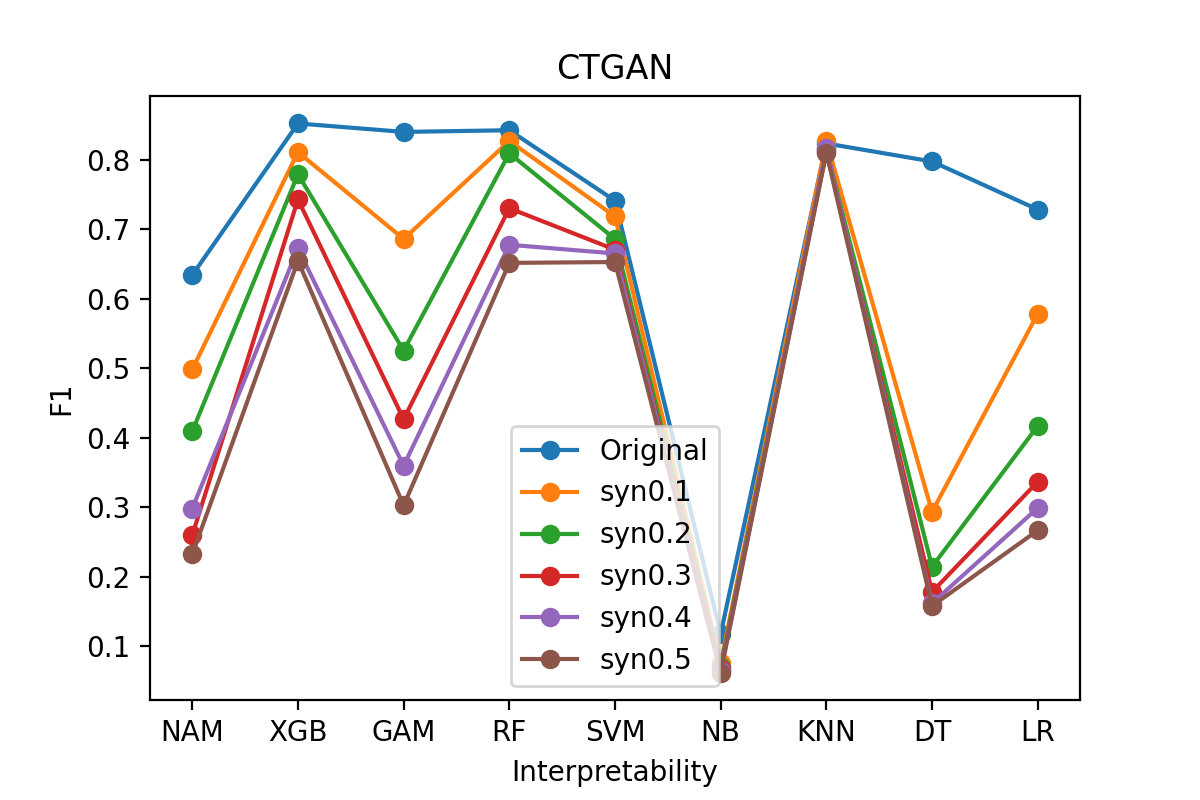}
    \includegraphics[width=.50\textwidth]{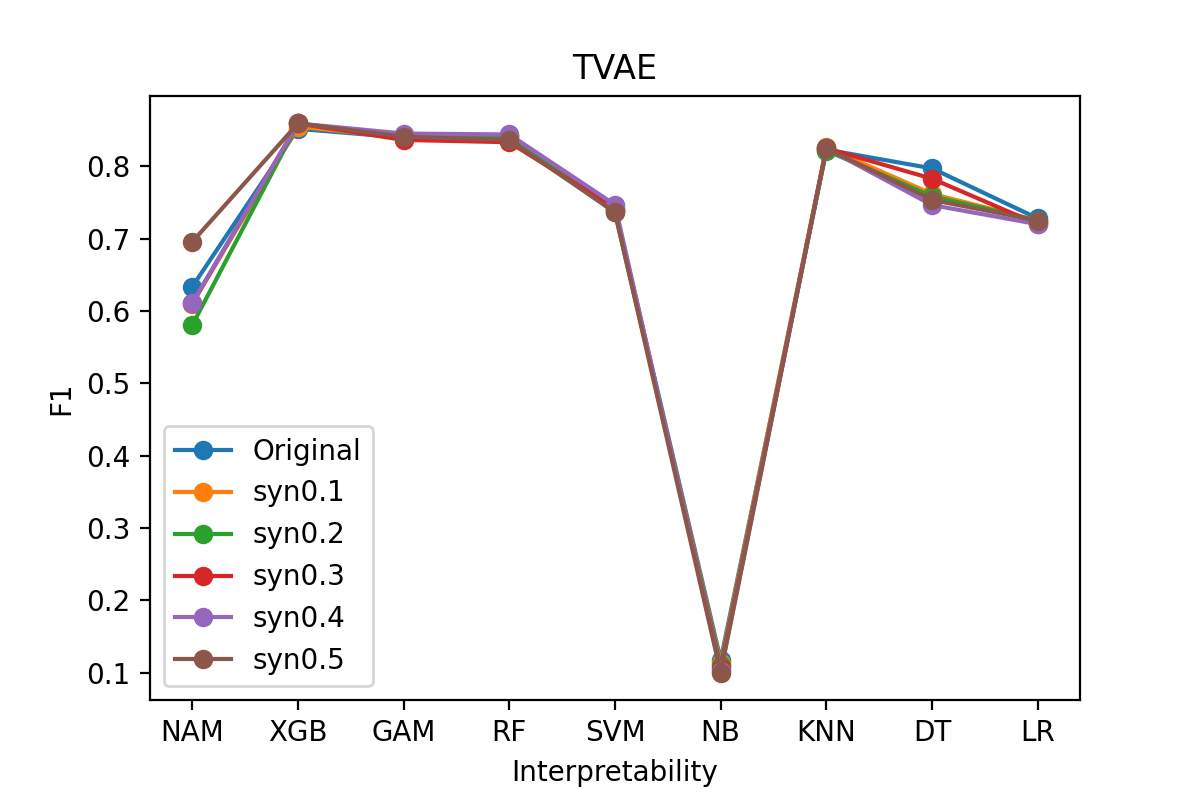}
    \includegraphics[width=.50\textwidth]{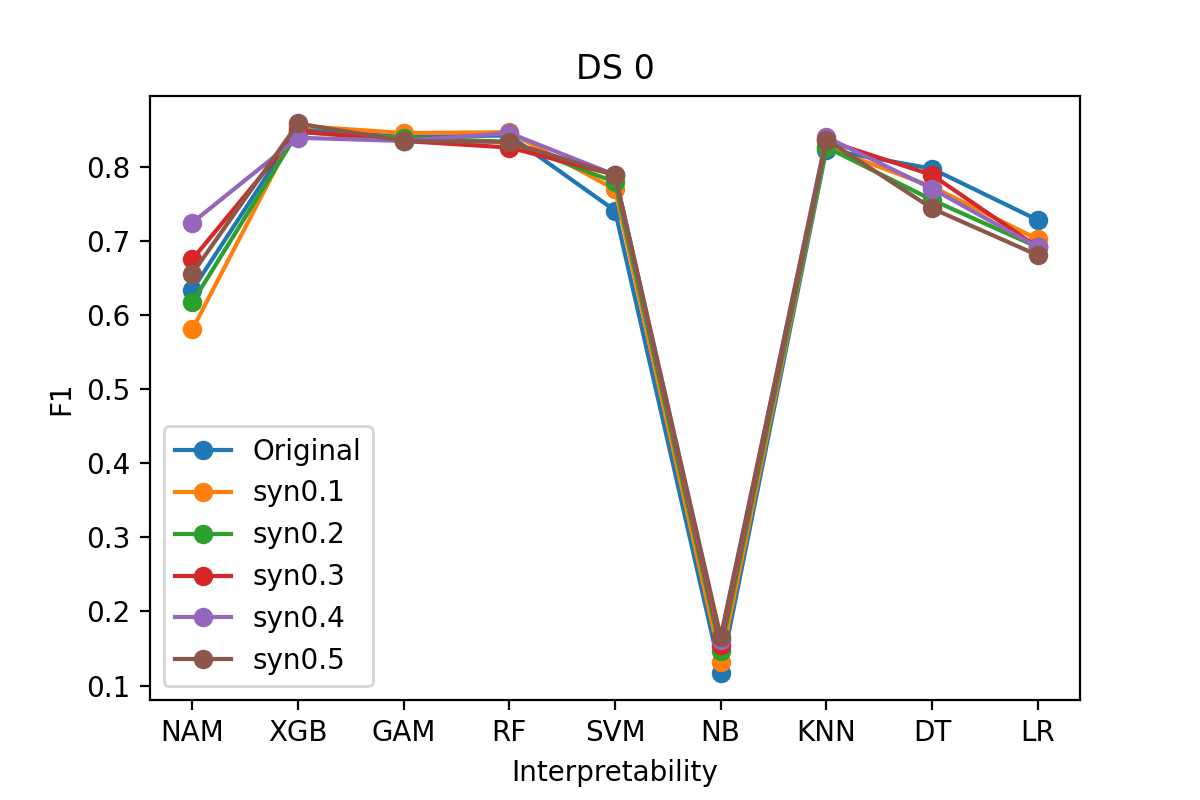}
    \includegraphics[width=.50\textwidth]{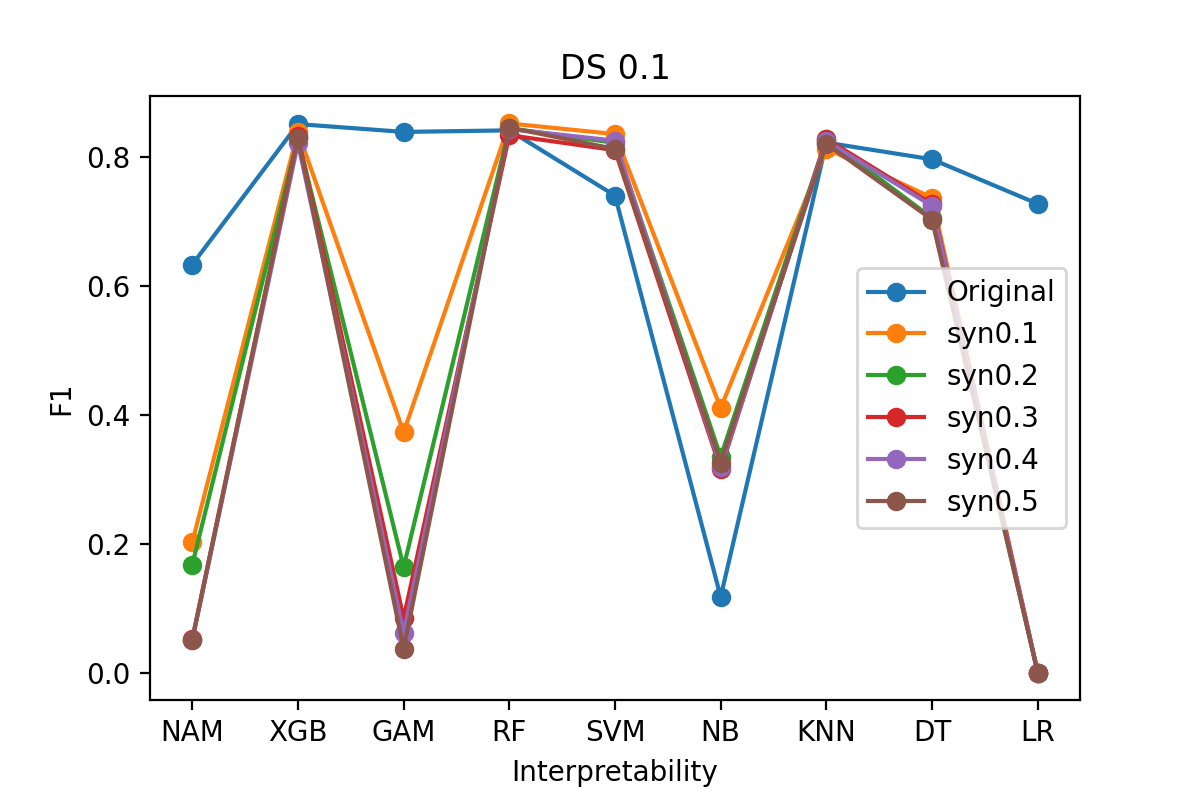}
    \caption{F1. Both CTGAN-augmented   and PrivBayes-augmented training dataset damages synthetic trained classifier utility.}
    \label{SynAug:f1}
\end{figure*}

\begin{figure*}[t]
    \includegraphics[width=.50\textwidth]{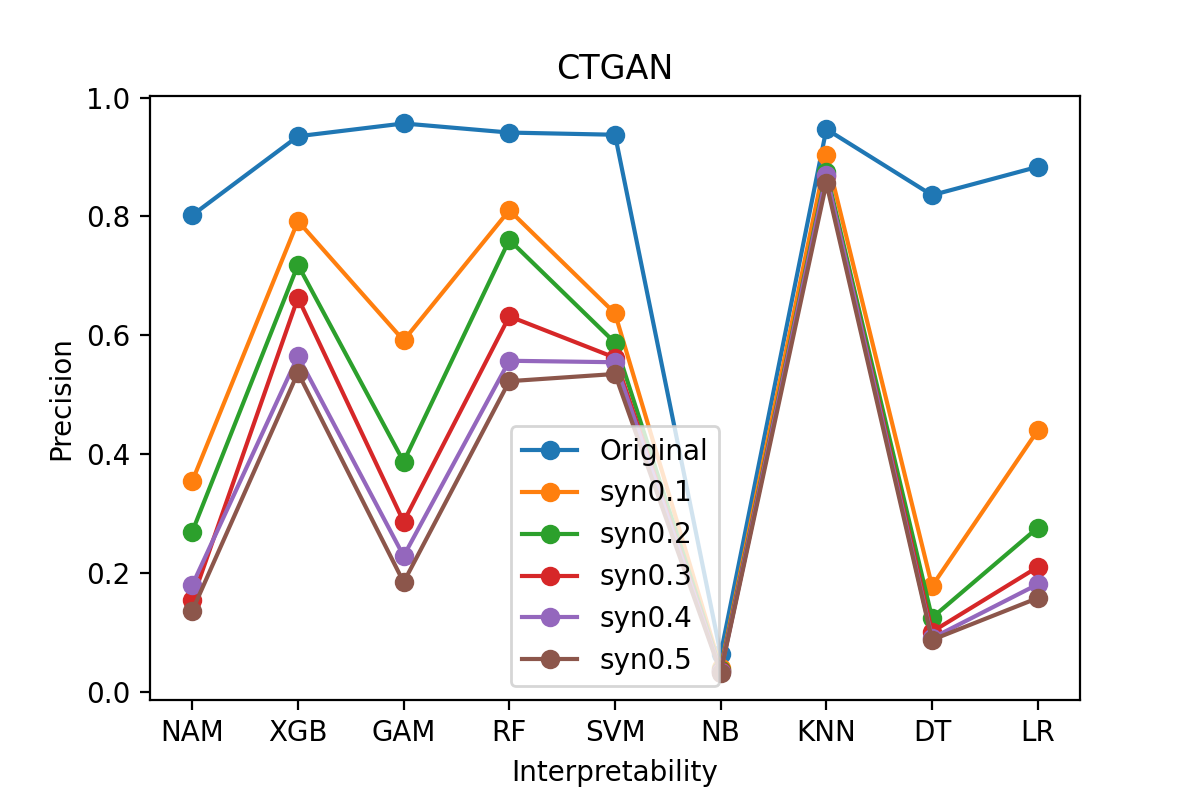}
    \includegraphics[width=.50\textwidth]{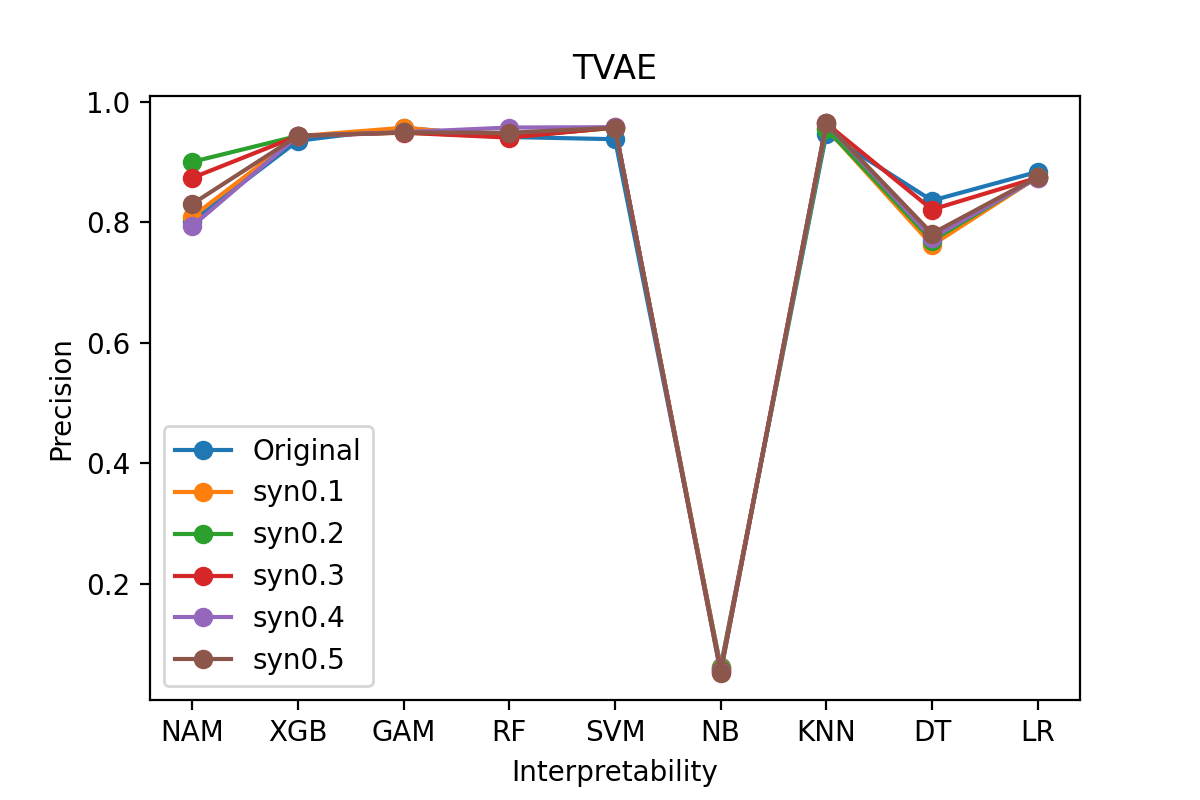}
    \includegraphics[width=.50\textwidth]{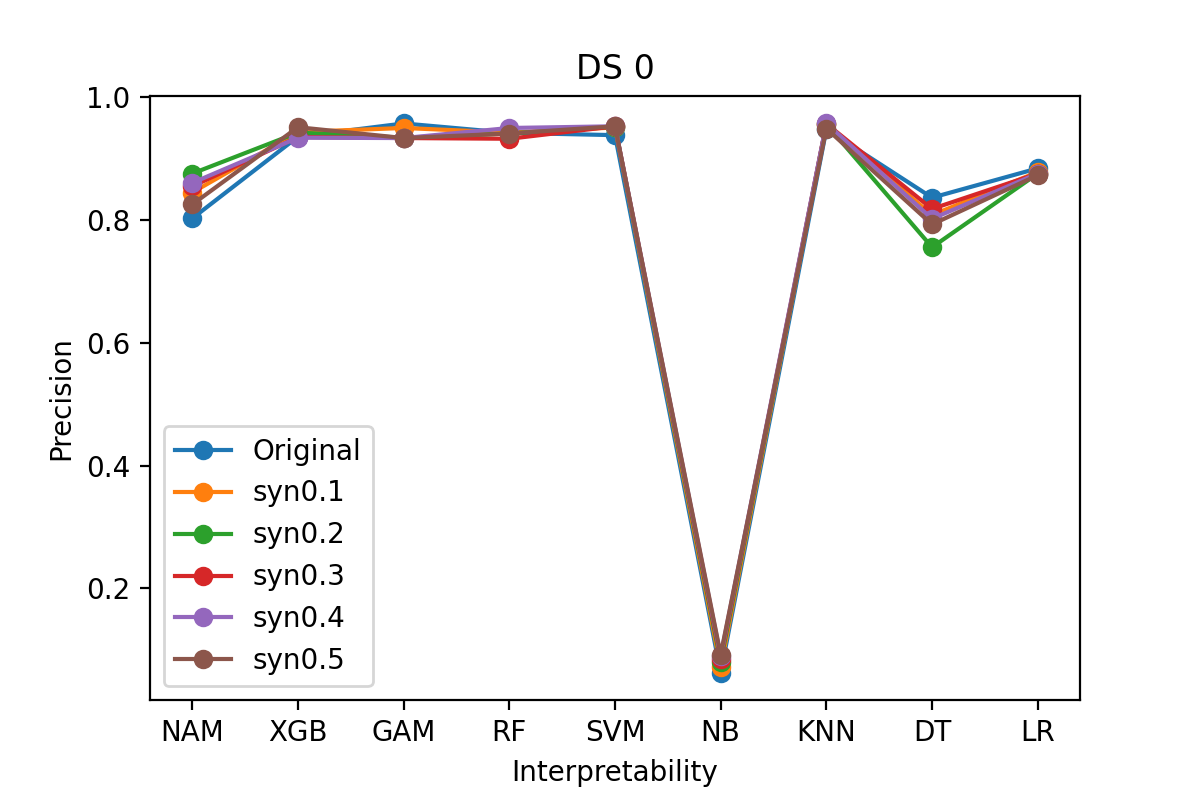}
    \includegraphics[width=.50\textwidth]{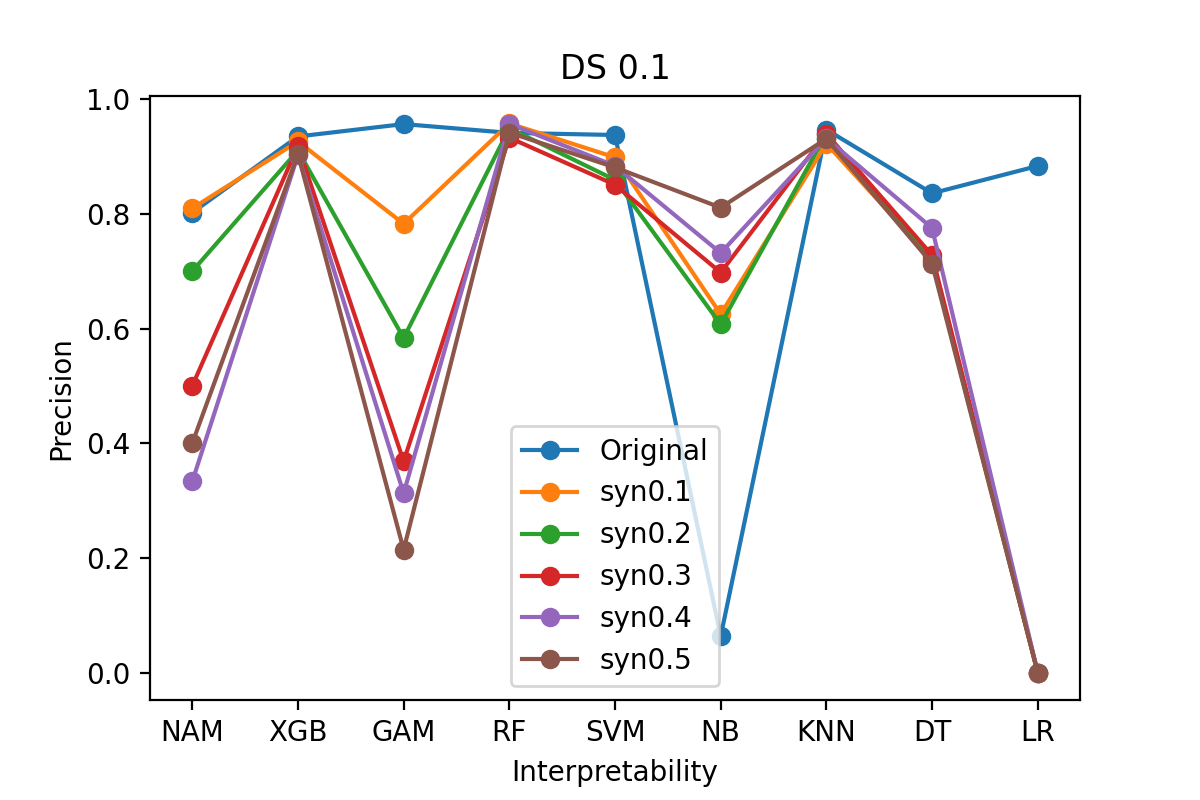}
    \caption{Precision. Both CTGAN-augmented and PrivBayes-augmented training dataset damages synthetic trained classifier utility.}
    \label{SynAug:precision}
\end{figure*}

\textbf{Metric-Oriented results.}
\begin{itemize}
    \item Figure \ref{SynAug:accuracy} gives Synthetic Augmented Training result under Accuracy performance metric. 
    \item Figure \ref{SynAug:auroc} gives Synthetic Augmented Training result under AUROC performance metric. 
    \item Figure \ref{SynAug:f1} gives Synthetic Augmented Training result under F1 performance metric. 
    \item Figure \ref{SynAug:recall} gives Synthetic Augmented Training result under Recall performance metric. 
    \item Figure \ref{SynAug:precision} gives Synthetic Augmented Training result under Precision performance metric. 
\end{itemize}

\textbf{Synthesizer-Oriented results.}

\begin{figure*}
    \includegraphics[width=.32\textwidth]{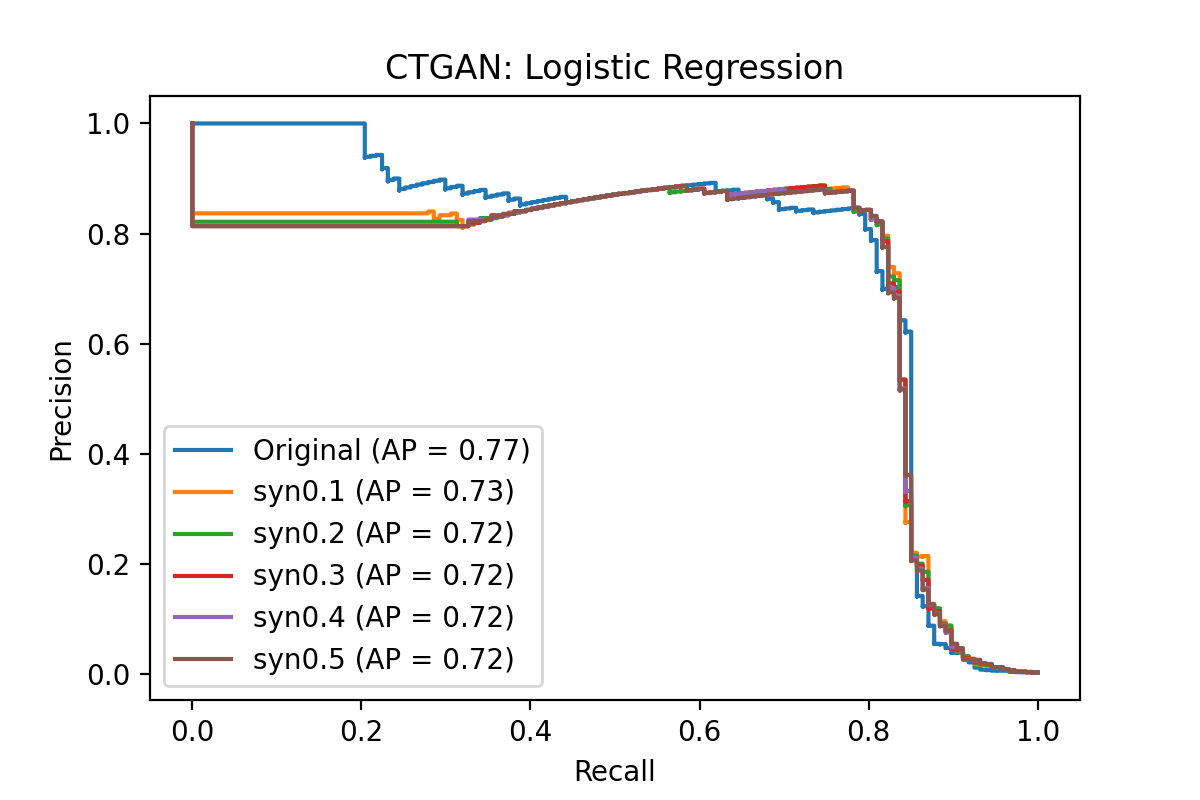}
    \includegraphics[width=.32\textwidth]{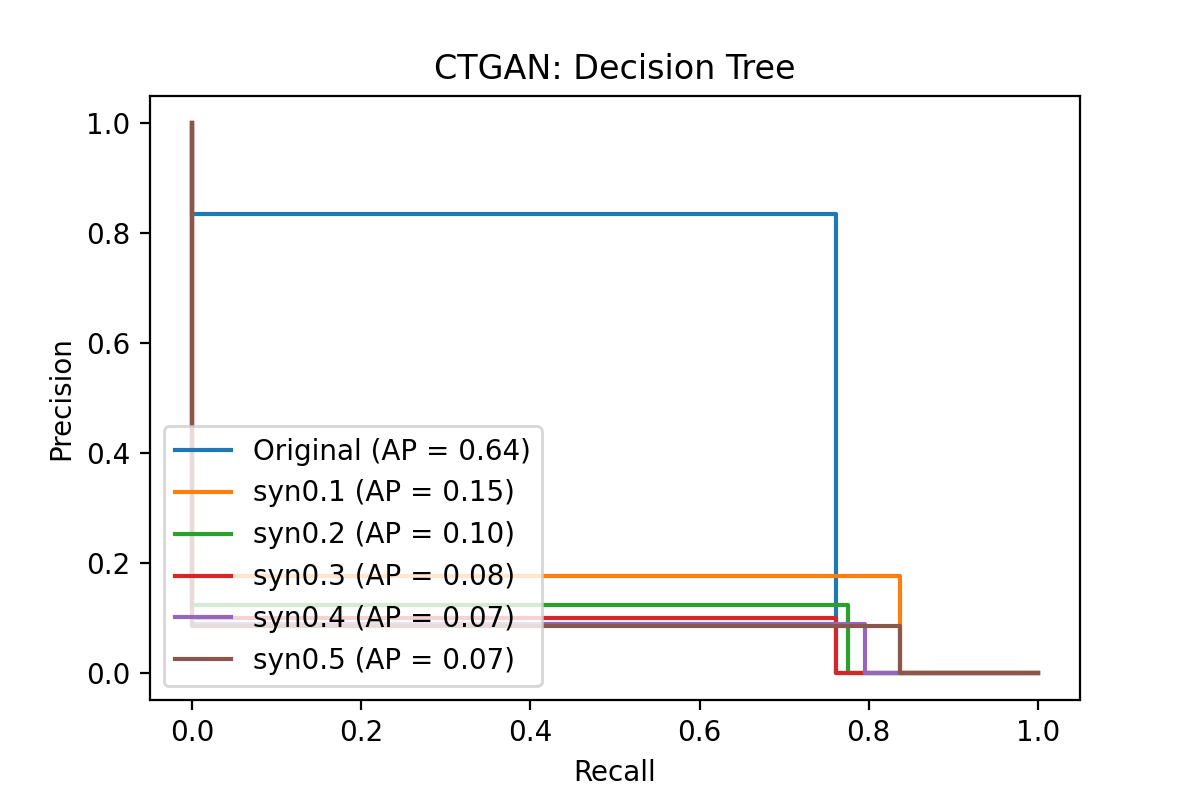}
    \includegraphics[width=.32\textwidth]{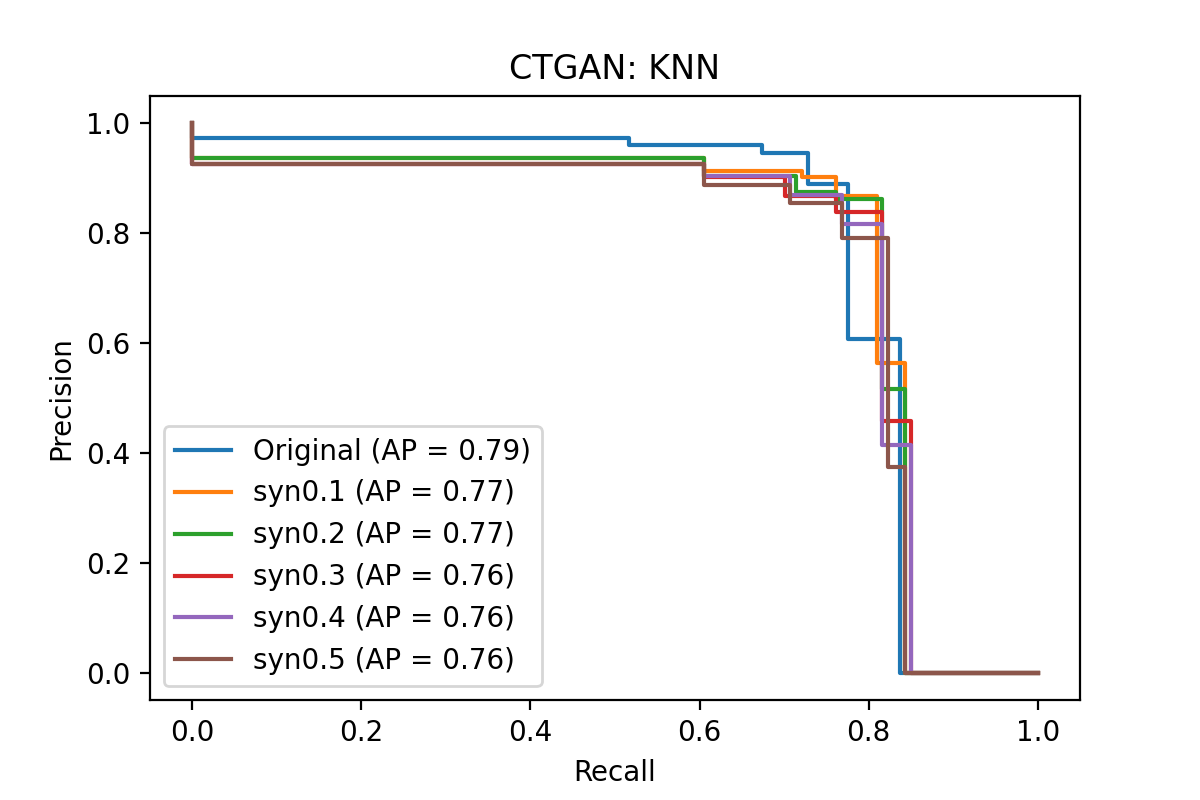}
    \includegraphics[width=.32\textwidth]{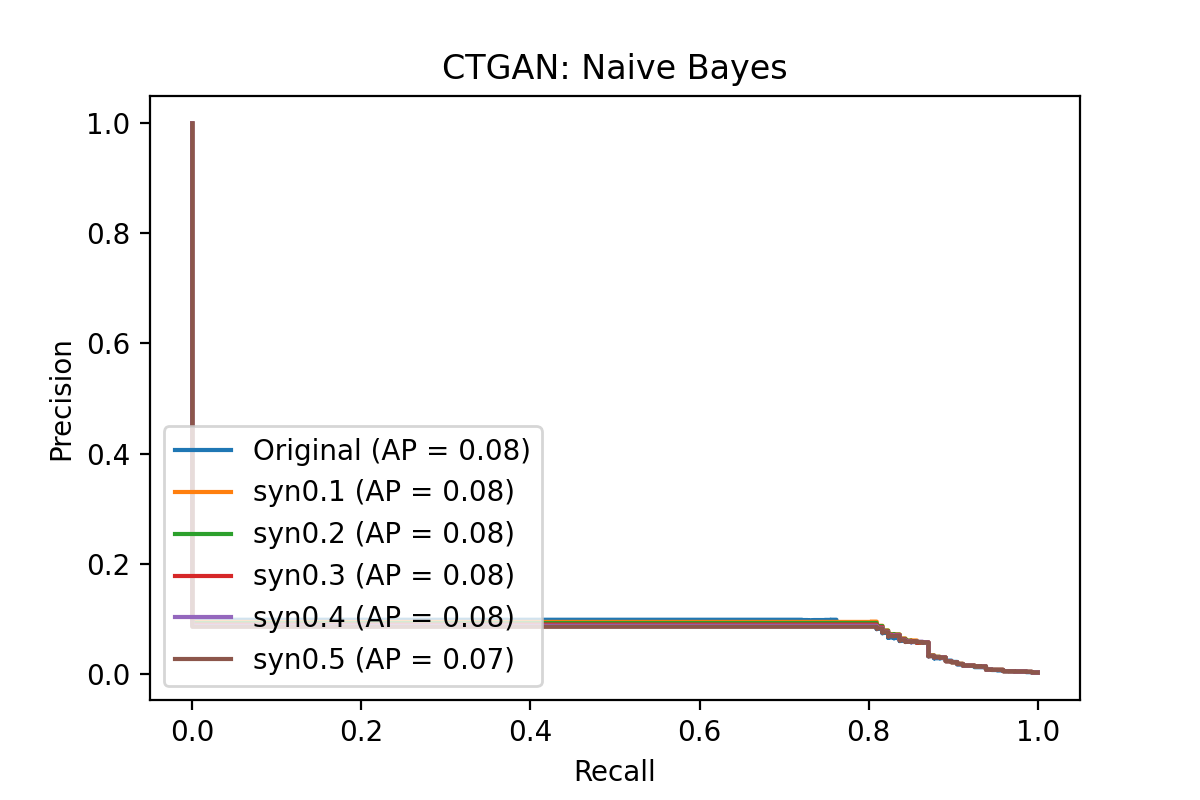}
    \includegraphics[width=.32\textwidth]{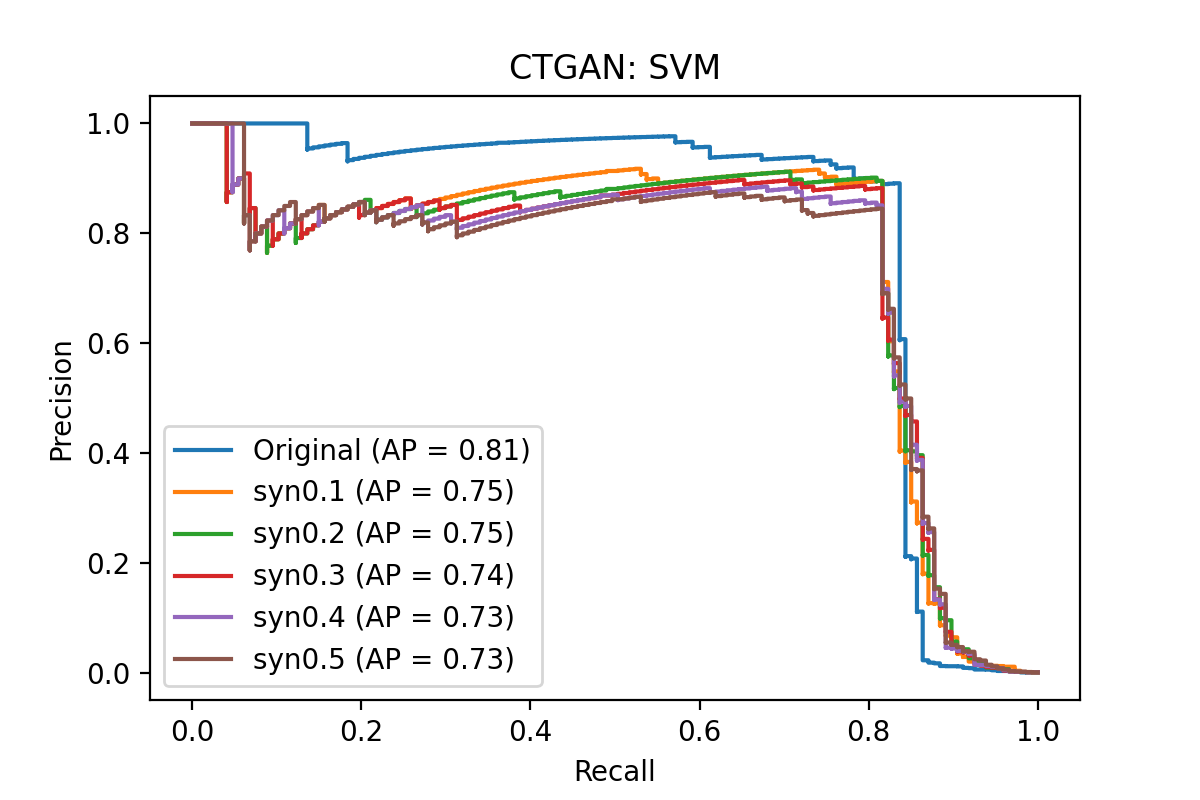}
    \includegraphics[width=.32\textwidth]{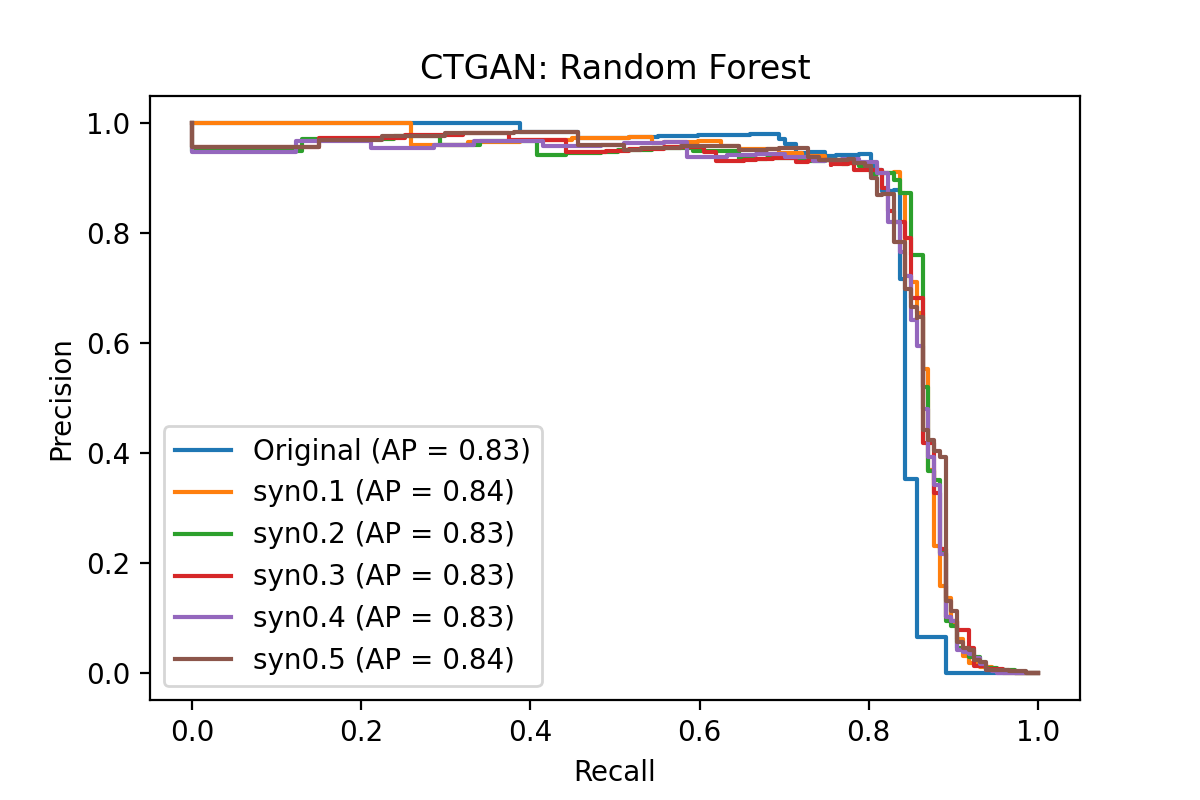}
    \includegraphics[width=.32\textwidth]{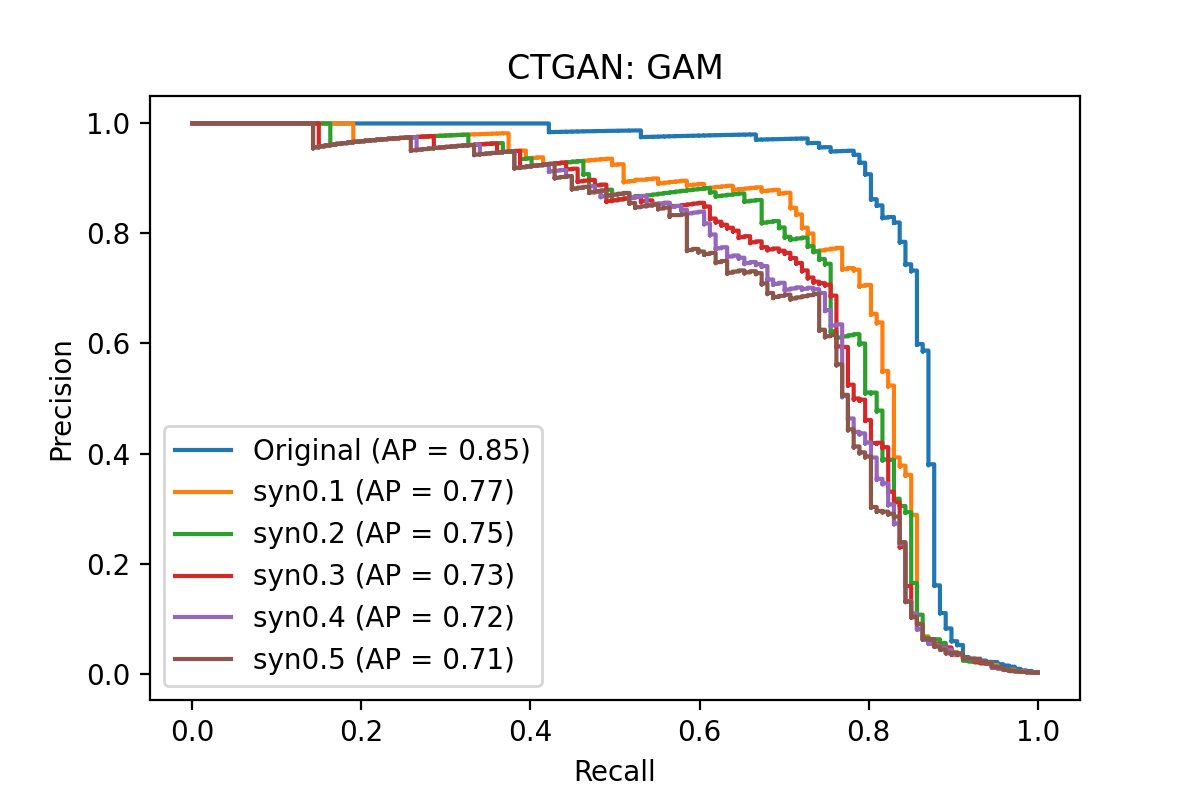}
    \includegraphics[width=.32\textwidth]{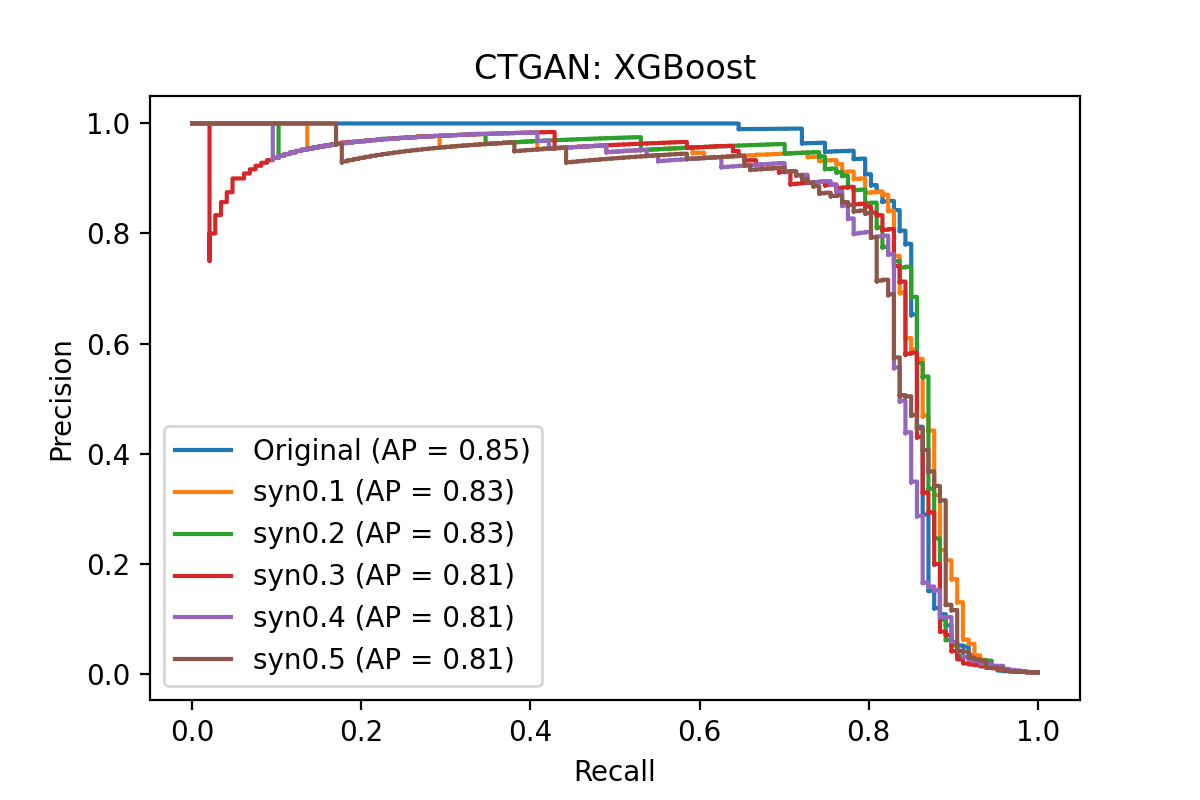}
    \includegraphics[width=.32\textwidth]{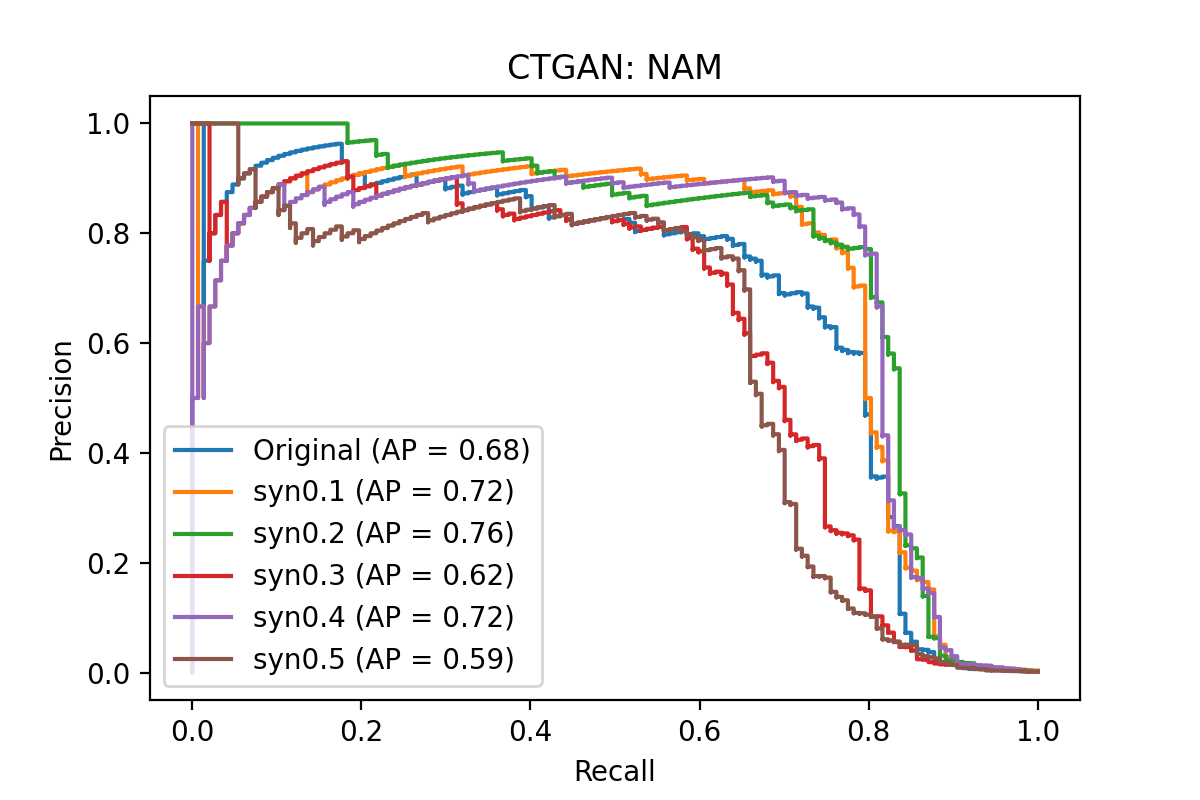}
    \caption{Precision-Recall curve for CTGAN. CTGAN-augmented training datasets in general damages the Precision-Recall curve across all ML classifier.}
    \label{SynAug:ctgan}
\end{figure*}

\begin{figure*}
    \includegraphics[width=.32\textwidth]{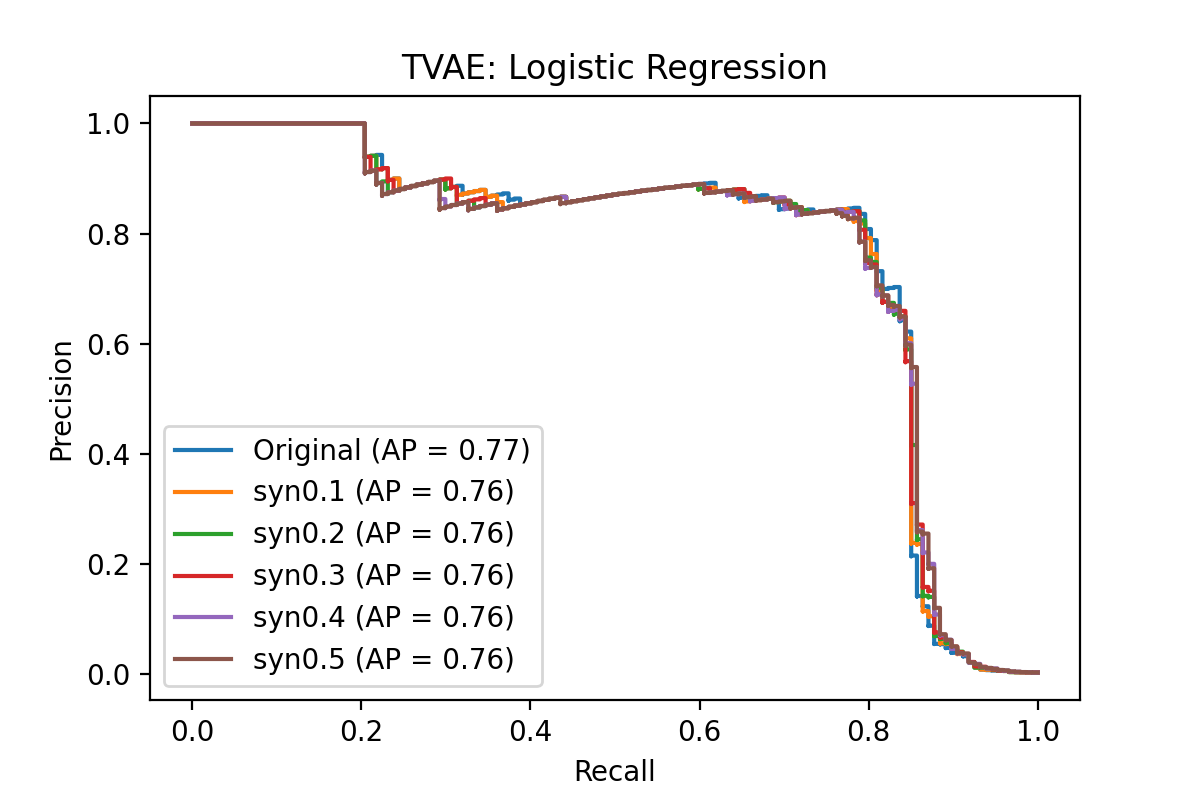}
    \includegraphics[width=.32\textwidth]{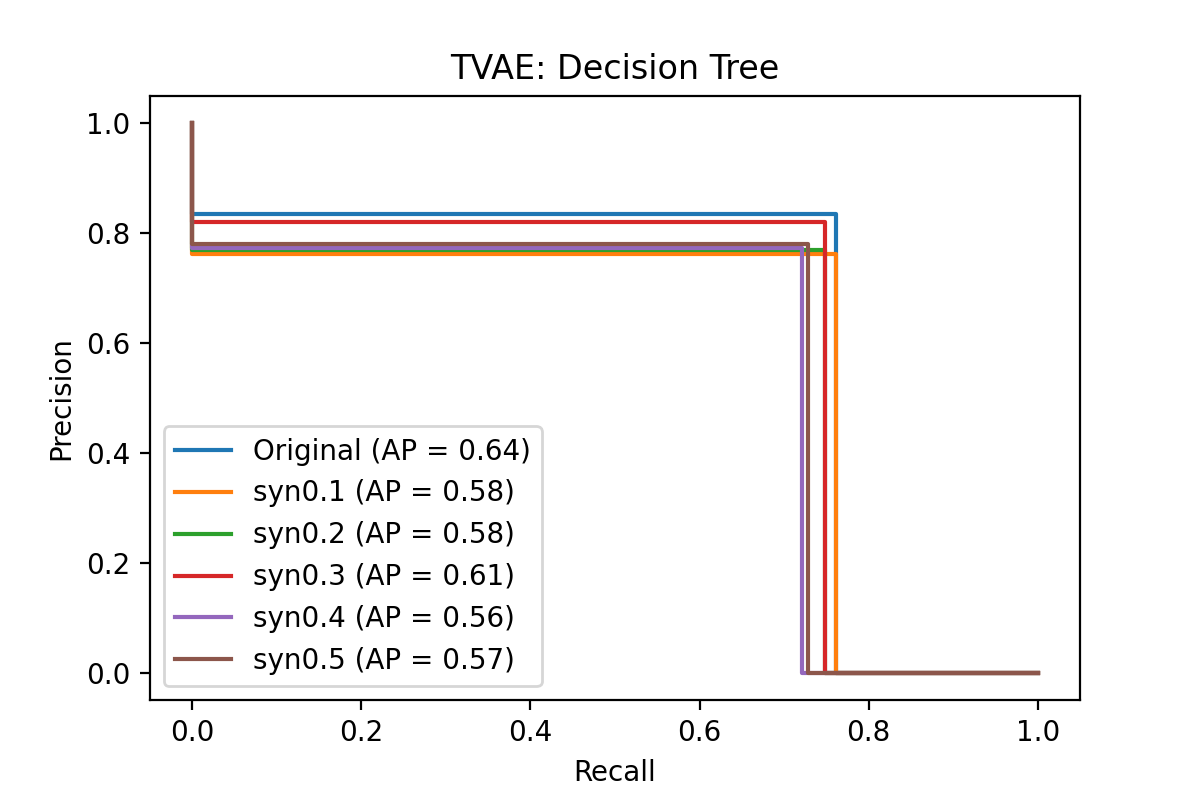}
    \includegraphics[width=.32\textwidth]{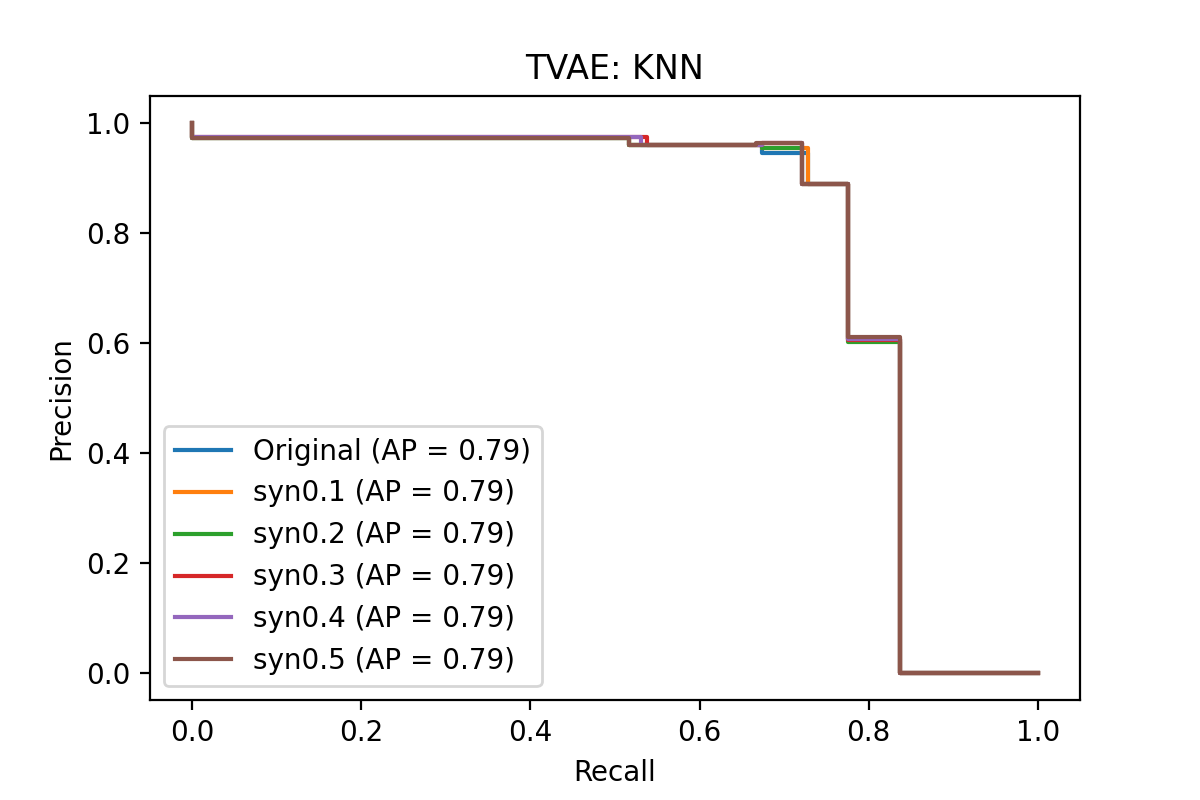}
    \includegraphics[width=.32\textwidth]{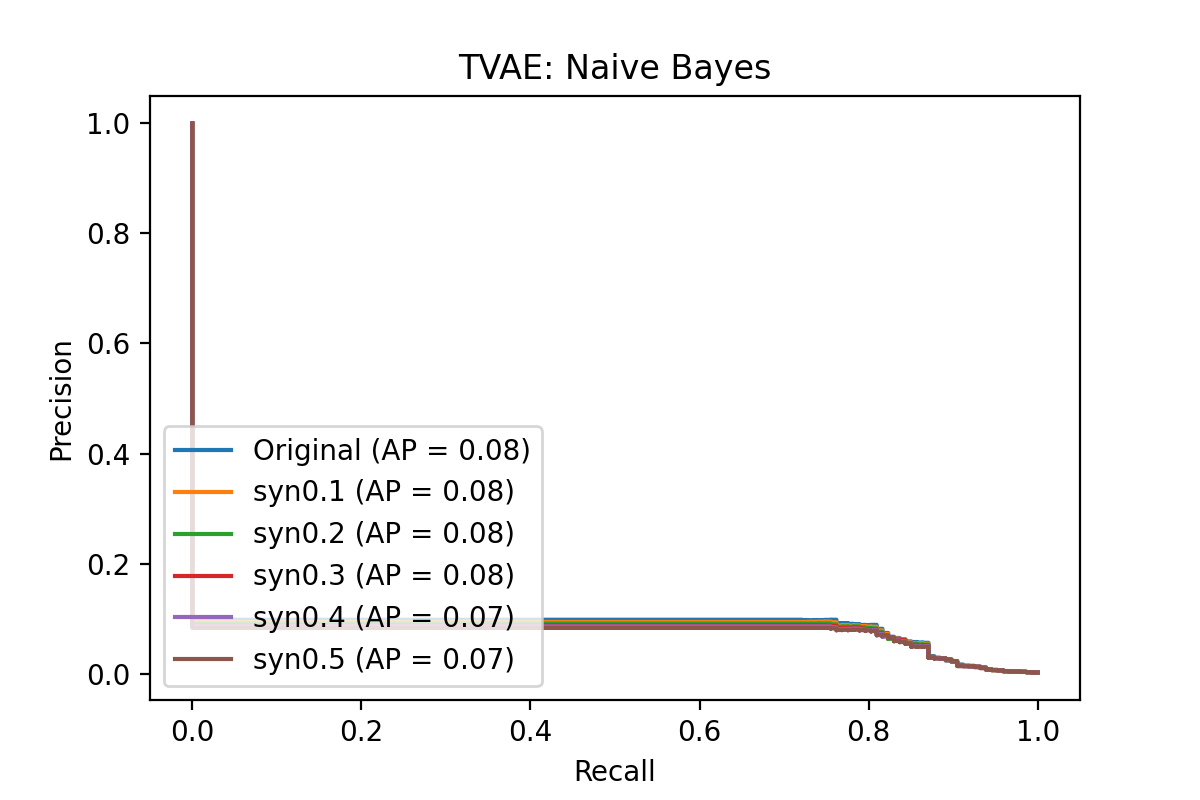}
    \includegraphics[width=.32\textwidth]{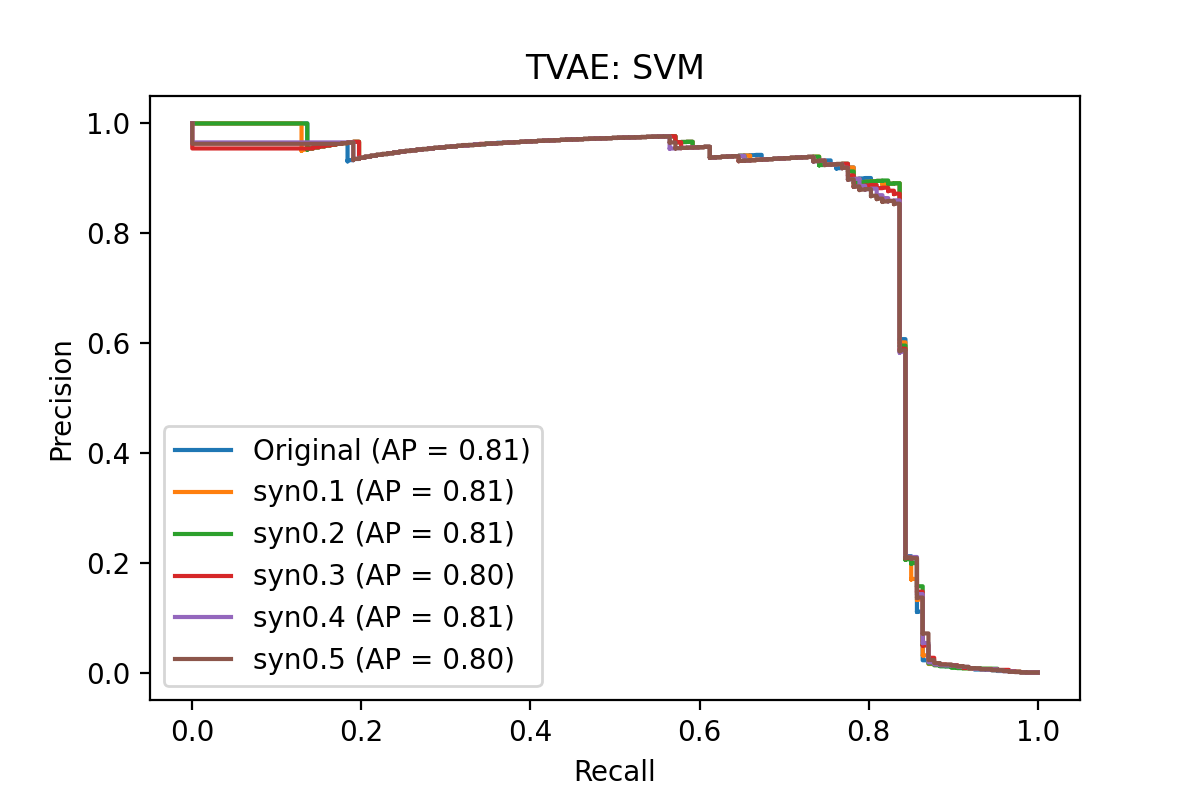}
    \includegraphics[width=.32\textwidth]{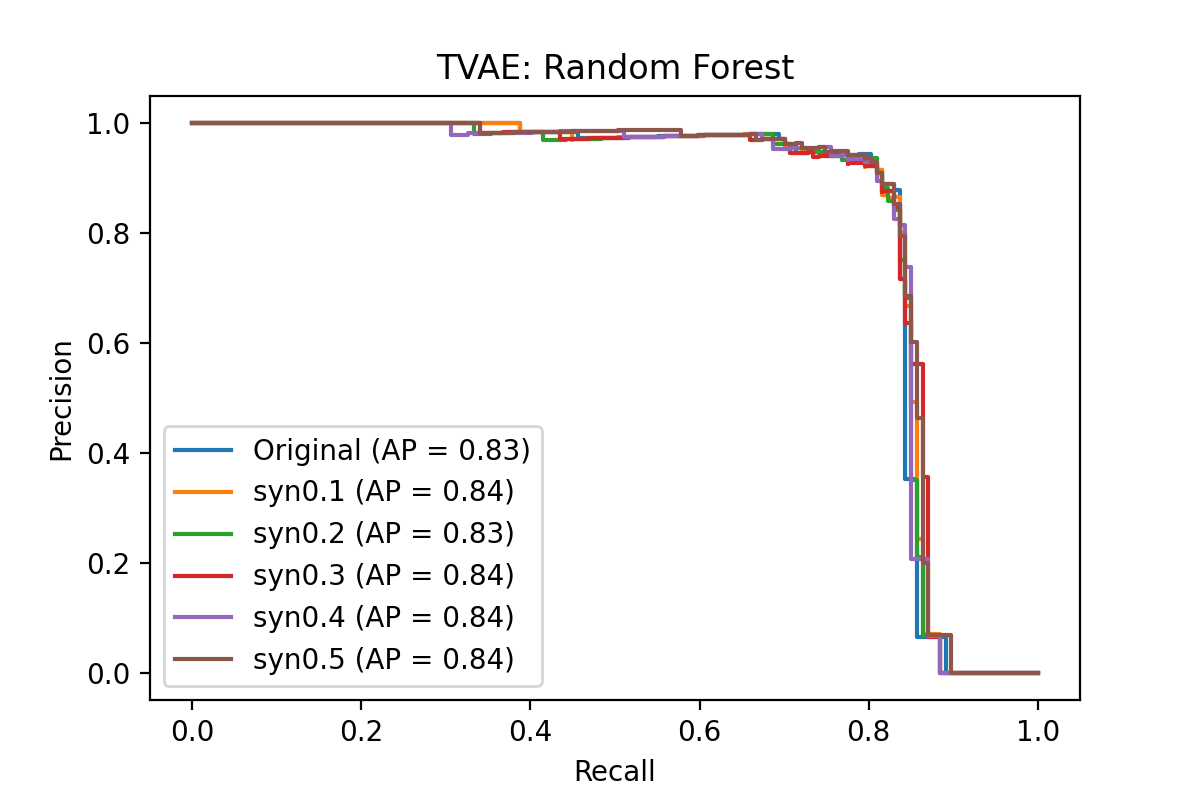}
    \includegraphics[width=.32\textwidth]{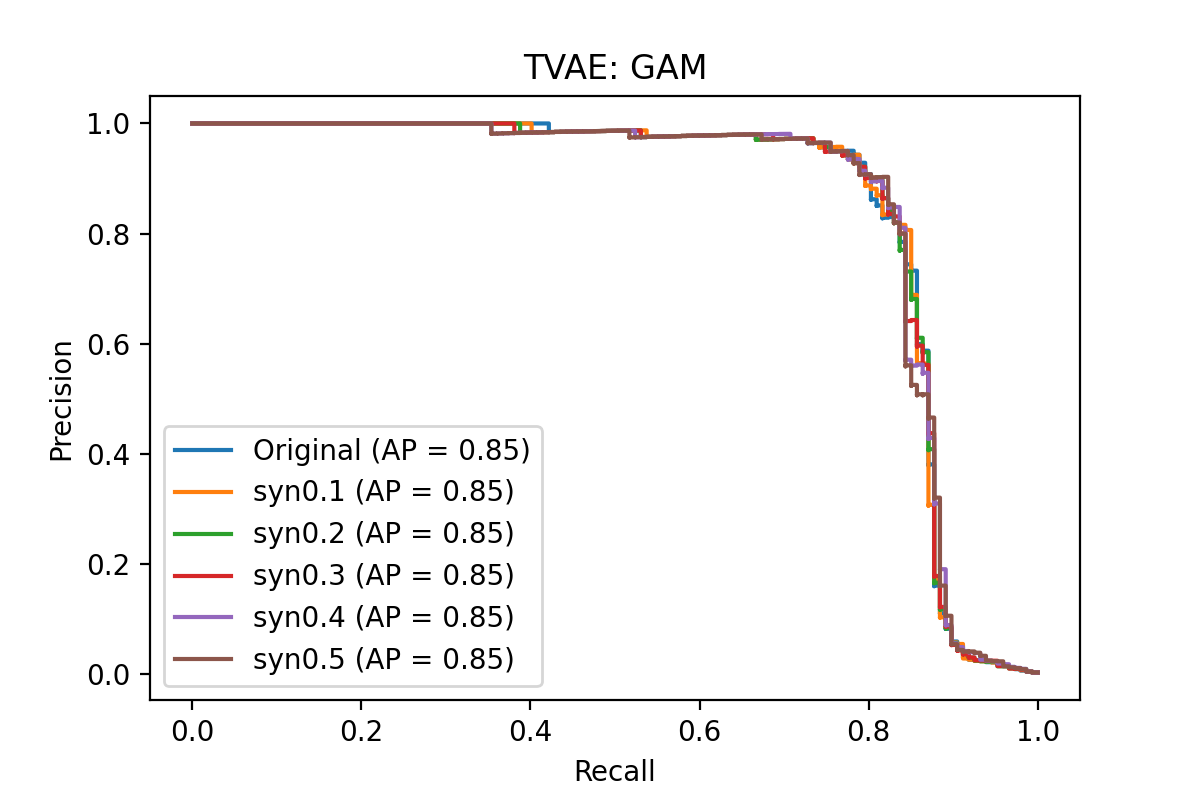}
    \includegraphics[width=.32\textwidth]{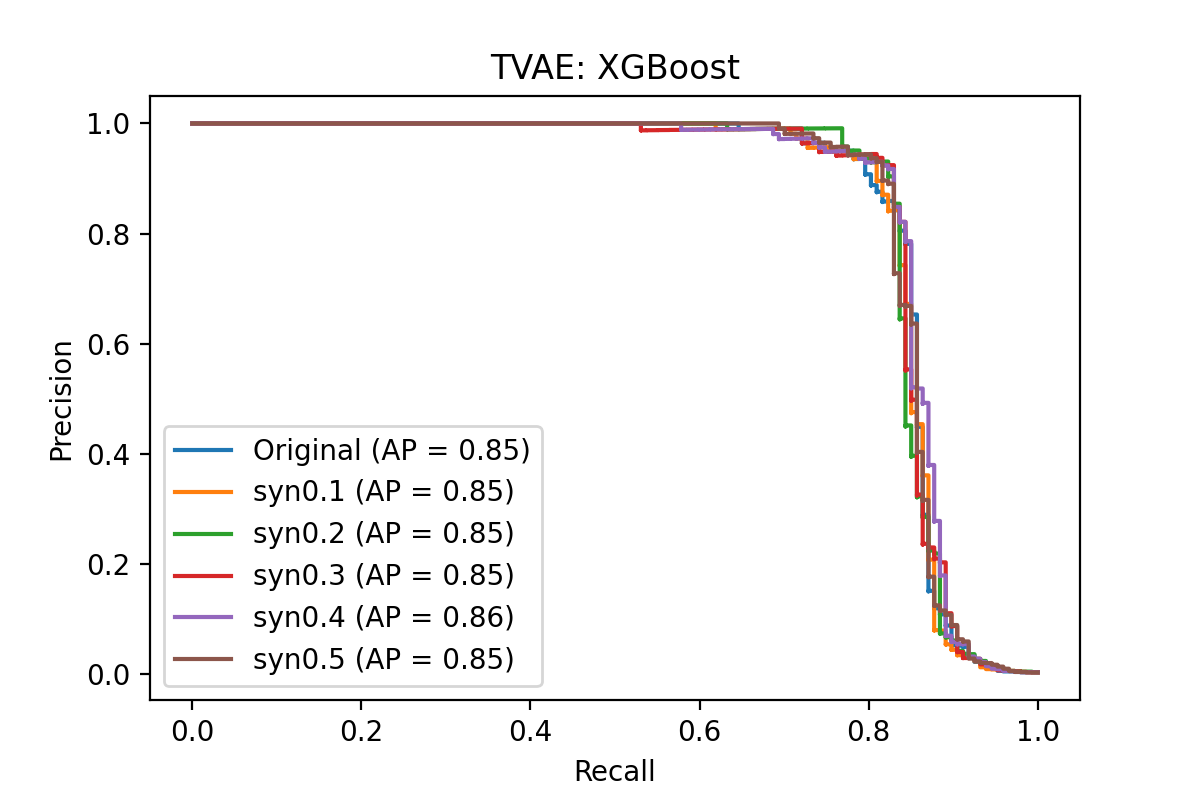}
    \includegraphics[width=.32\textwidth]{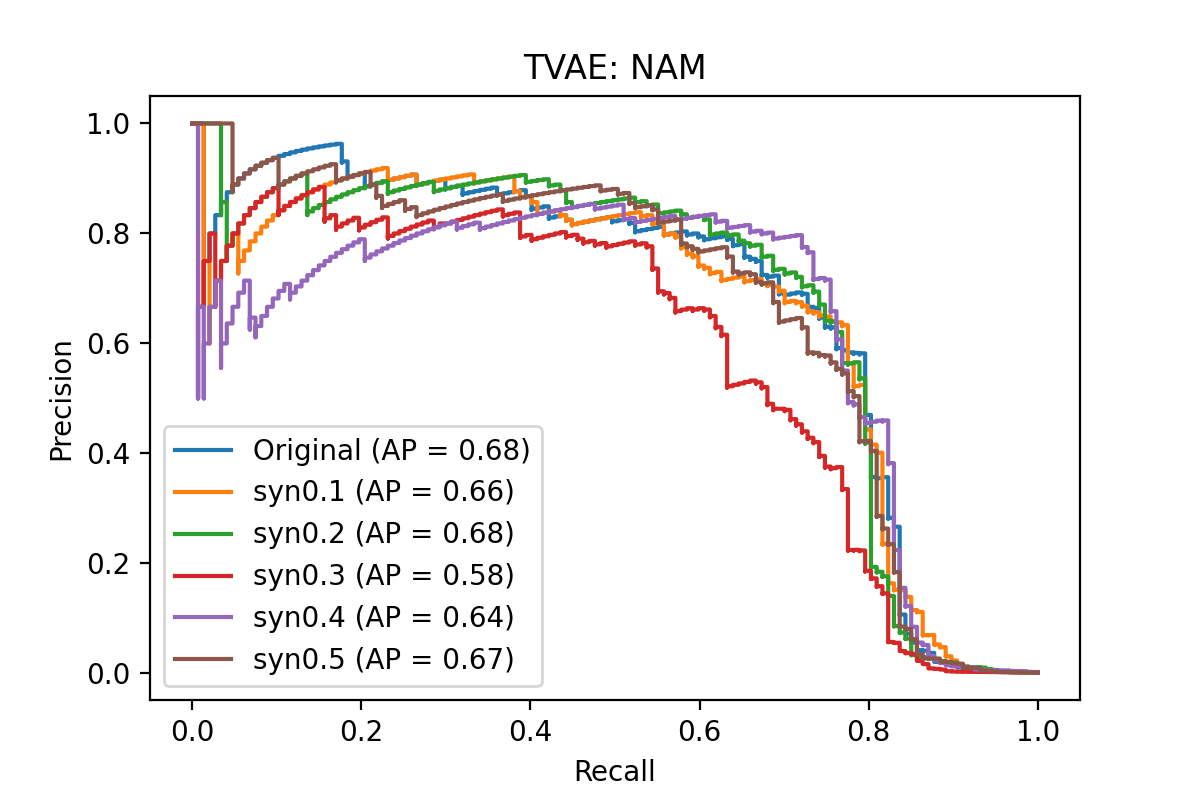}
    \caption{Precision-Recall curve for TVAE. TVAE-augmented training datasets in general do not improve or damage the Precision-Recall curve across all ML classifier.}
    \label{SynAug:tvae}
\end{figure*}

\begin{figure*}
    \includegraphics[width=.32\textwidth]{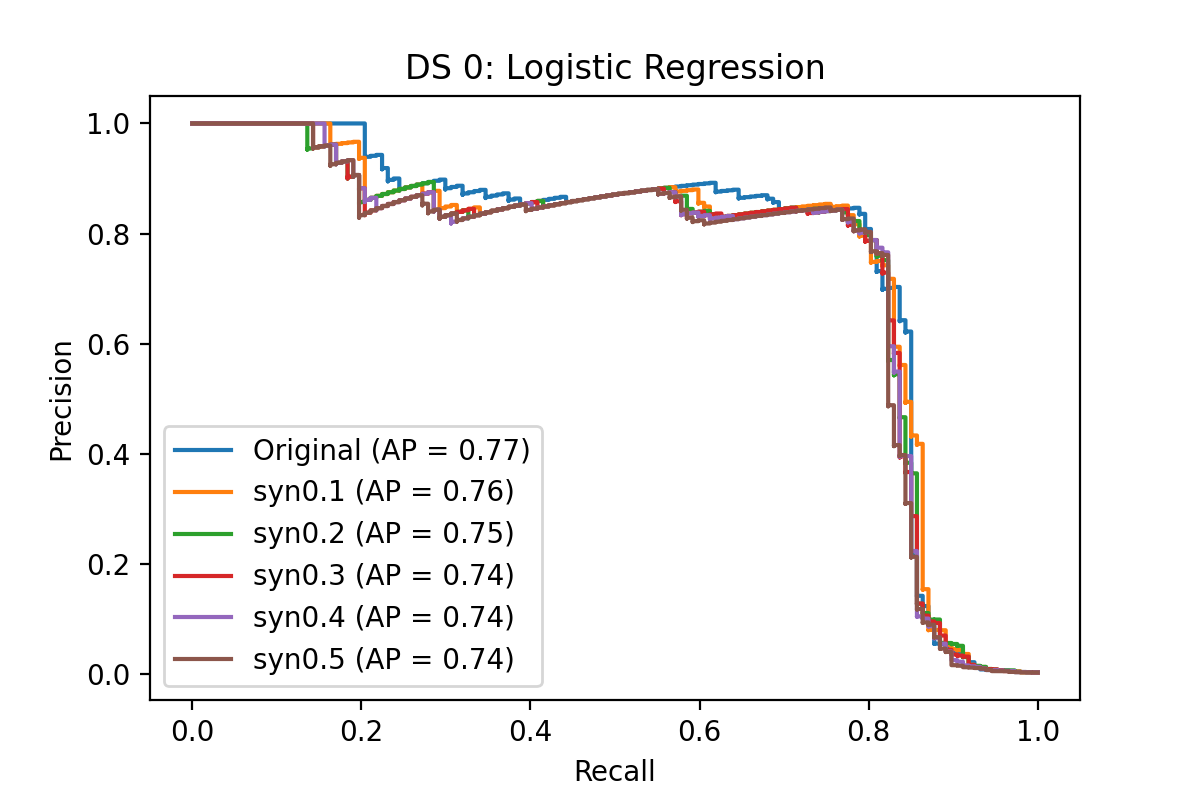}
    \includegraphics[width=.32\textwidth]{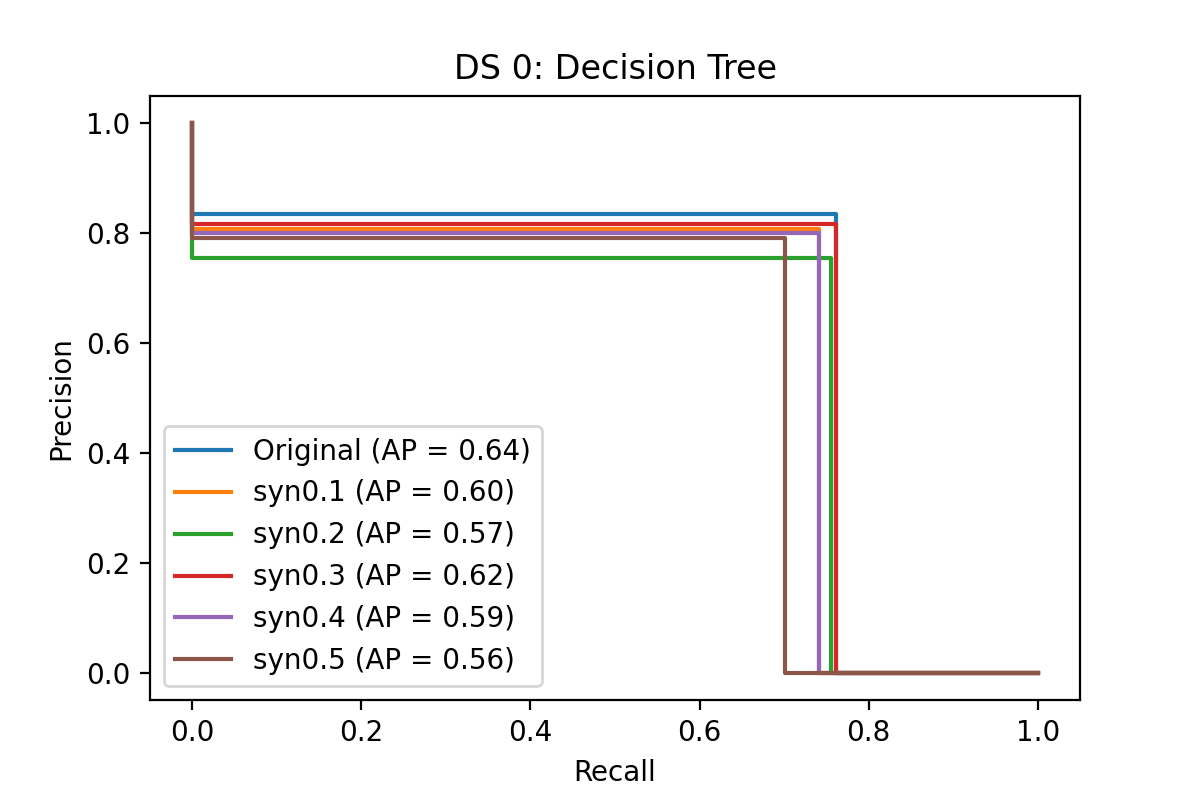}
    \includegraphics[width=.32\textwidth]{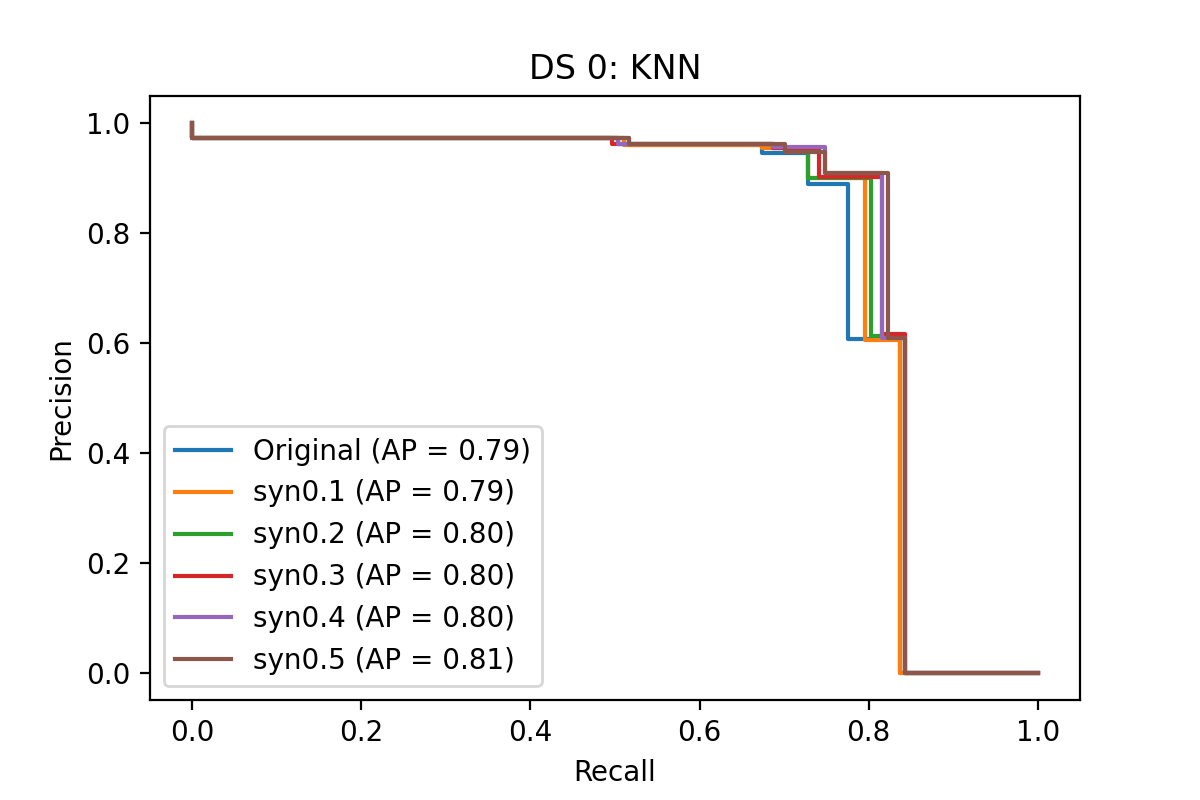}
    \includegraphics[width=.32\textwidth]{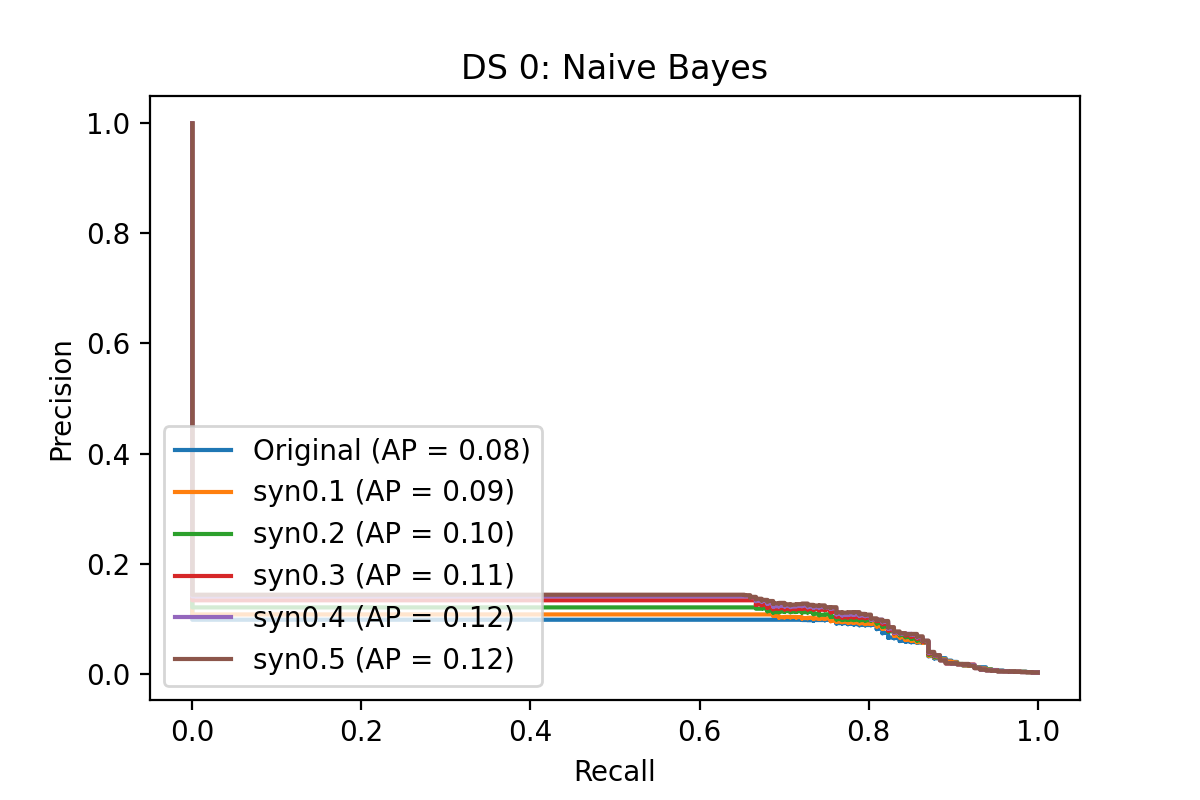}
    \includegraphics[width=.32\textwidth]{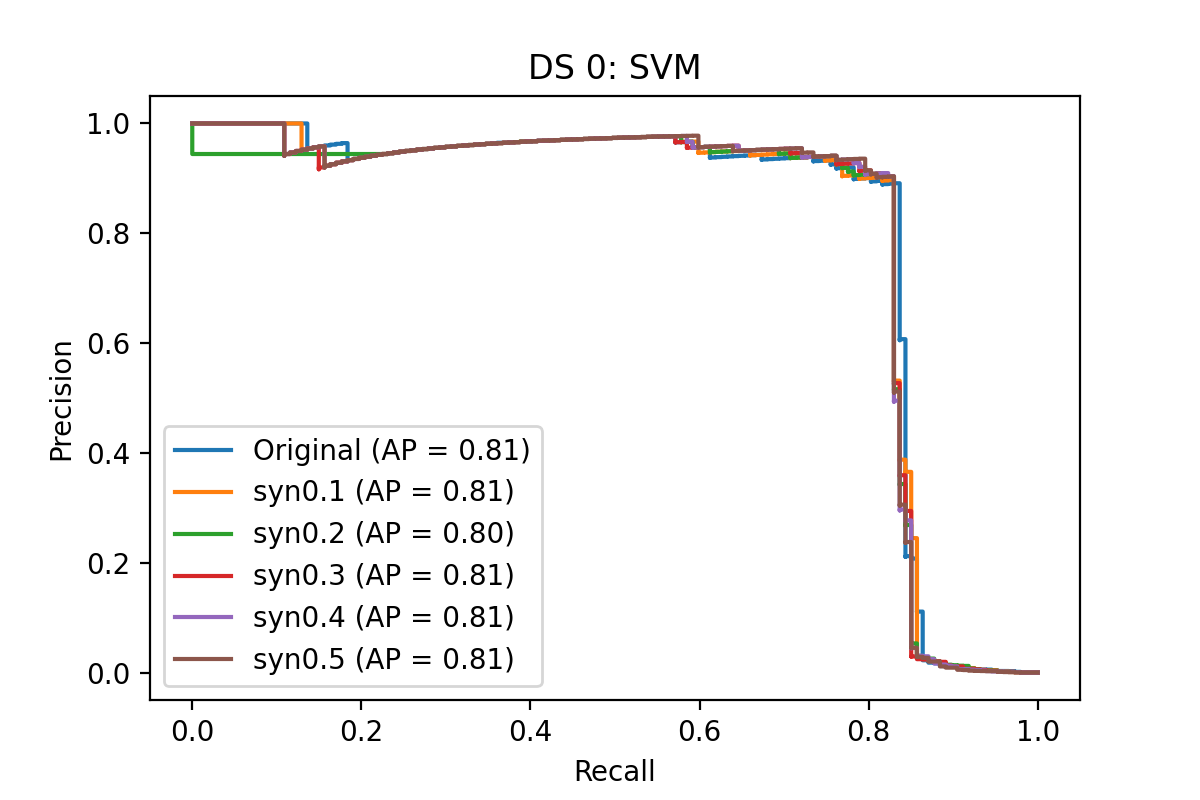}
    \includegraphics[width=.32\textwidth]{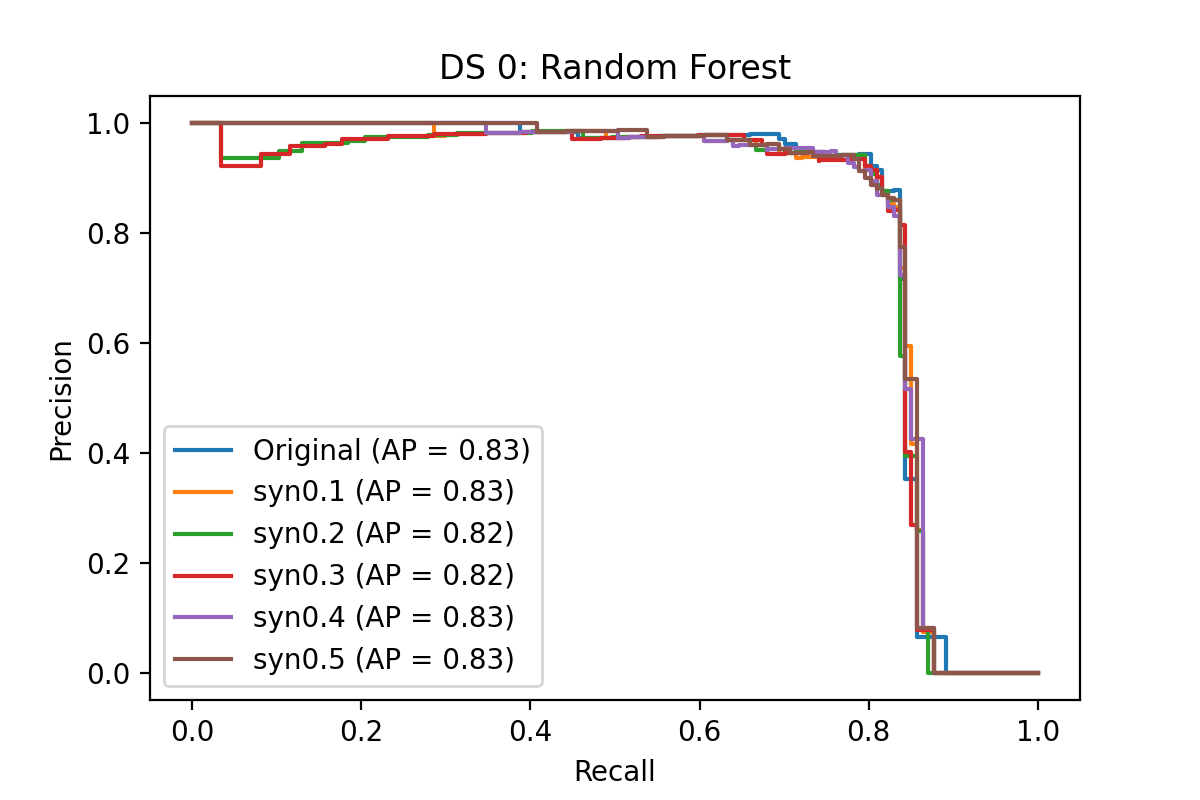}
    \includegraphics[width=.32\textwidth]{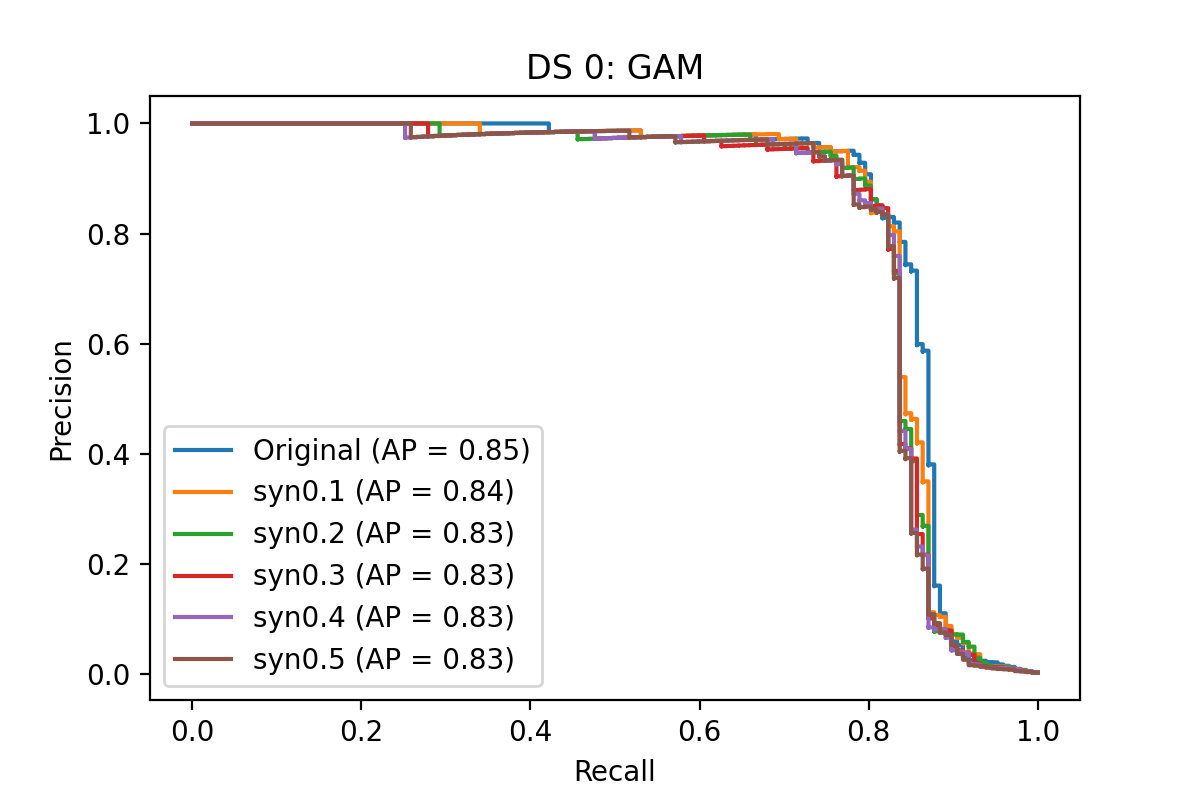}
    \includegraphics[width=.32\textwidth]{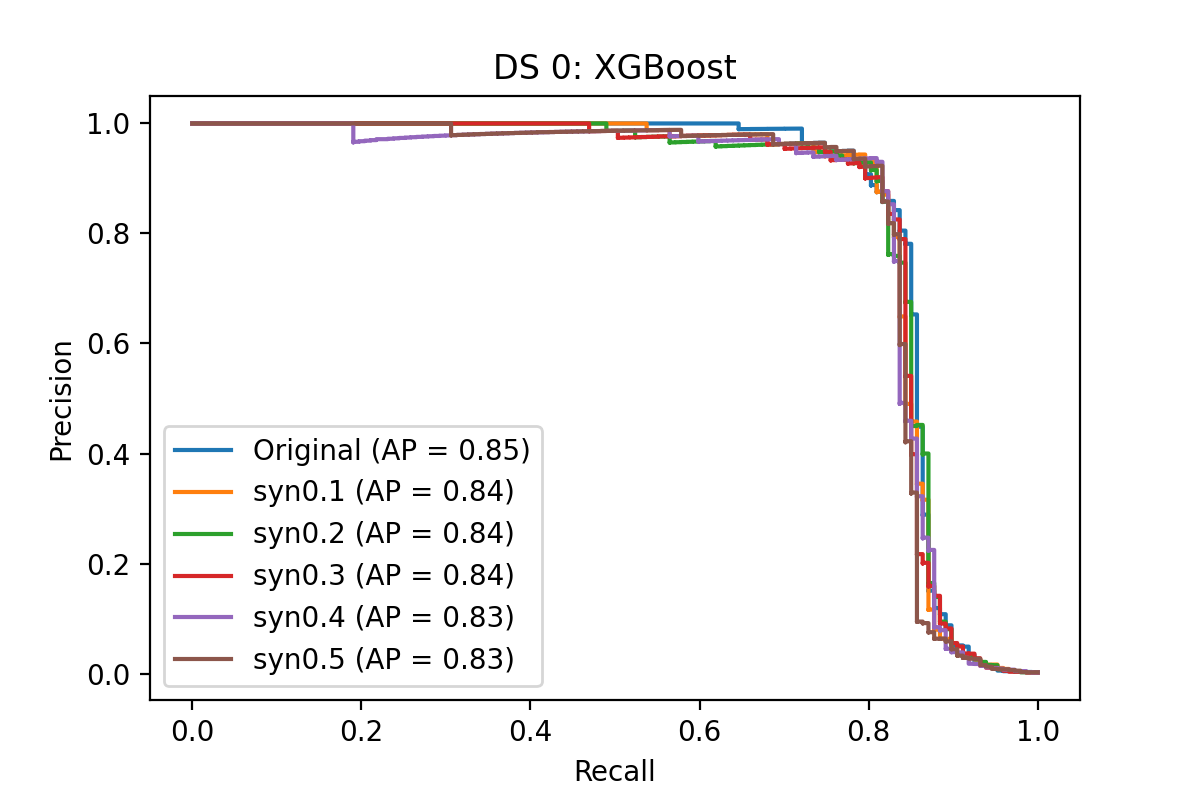}
    \includegraphics[width=.32\textwidth]{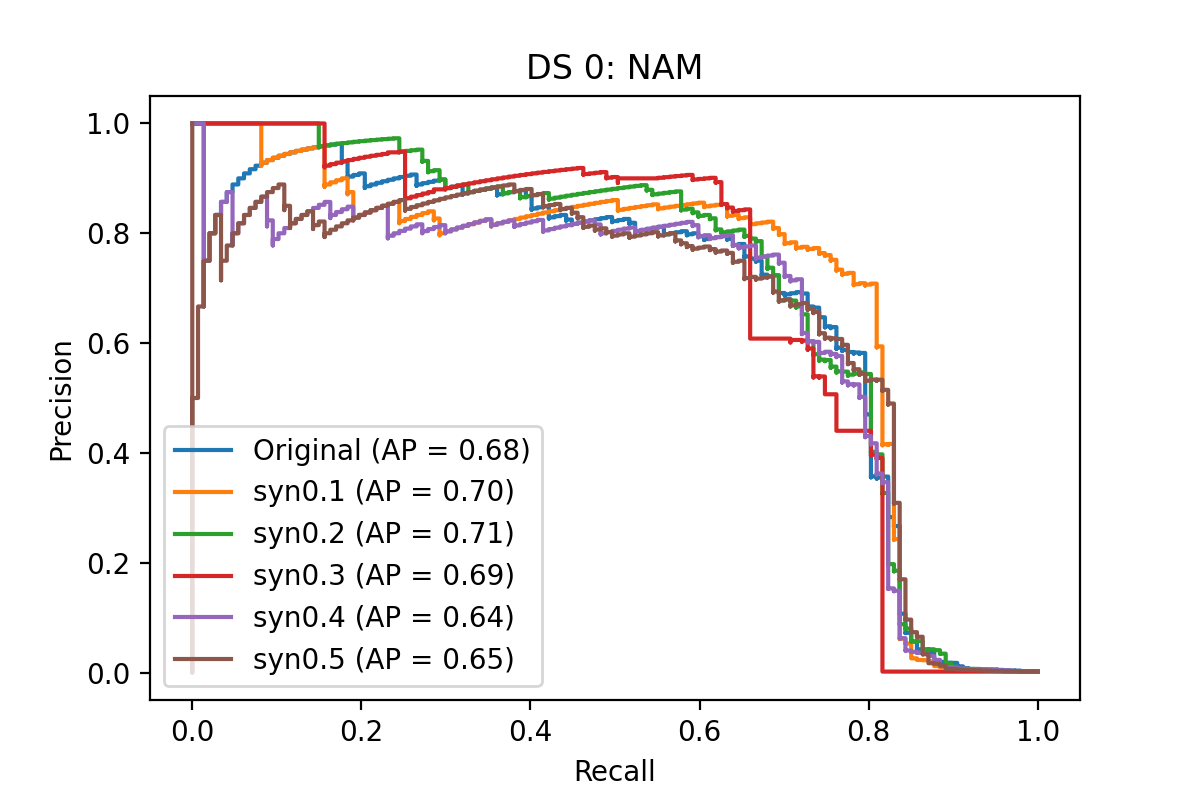}
    \caption{Precision-Recall curve for DS 0. DS 0-augmented training datasets in general do not improve or damage the Precision-Recall curve across all ML classifier.}
    \label{SynAug:ds0}
\end{figure*}

\begin{figure*}
    \includegraphics[width=.32\textwidth]{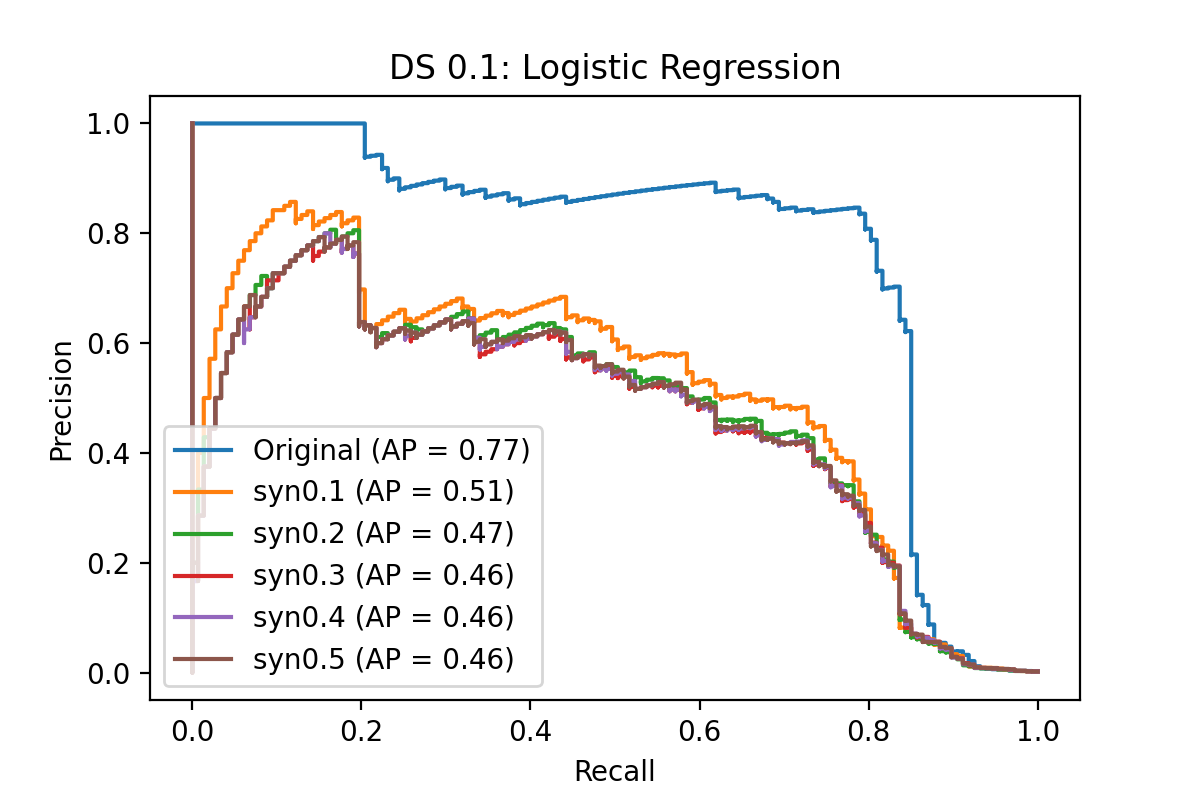}
    \includegraphics[width=.32\textwidth]{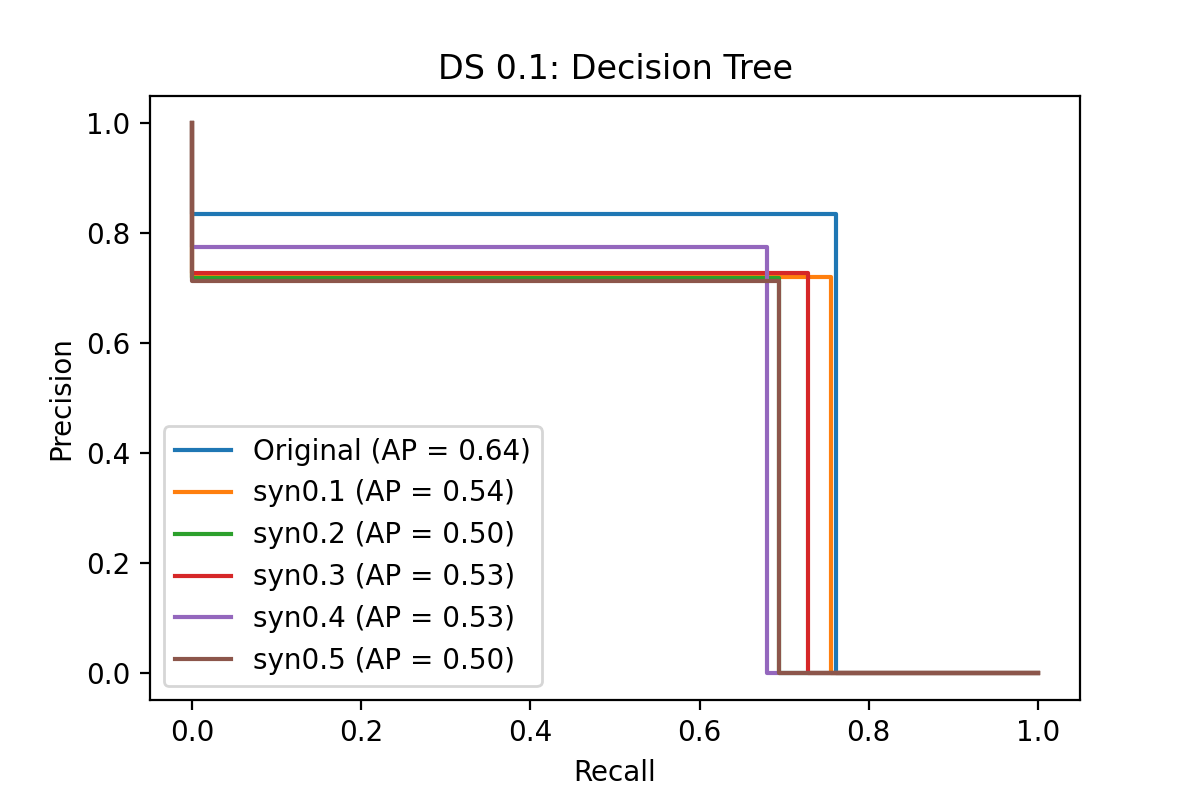}
    \includegraphics[width=.32\textwidth]{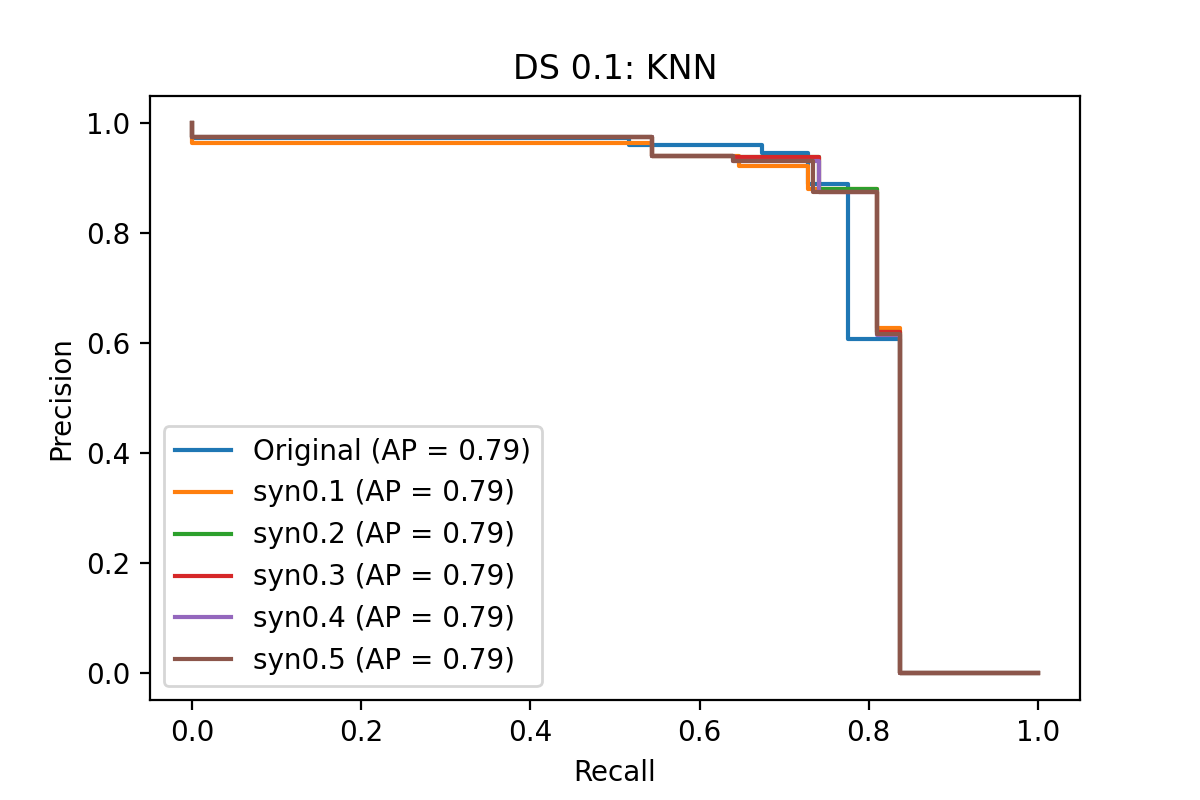}
    \includegraphics[width=.32\textwidth]{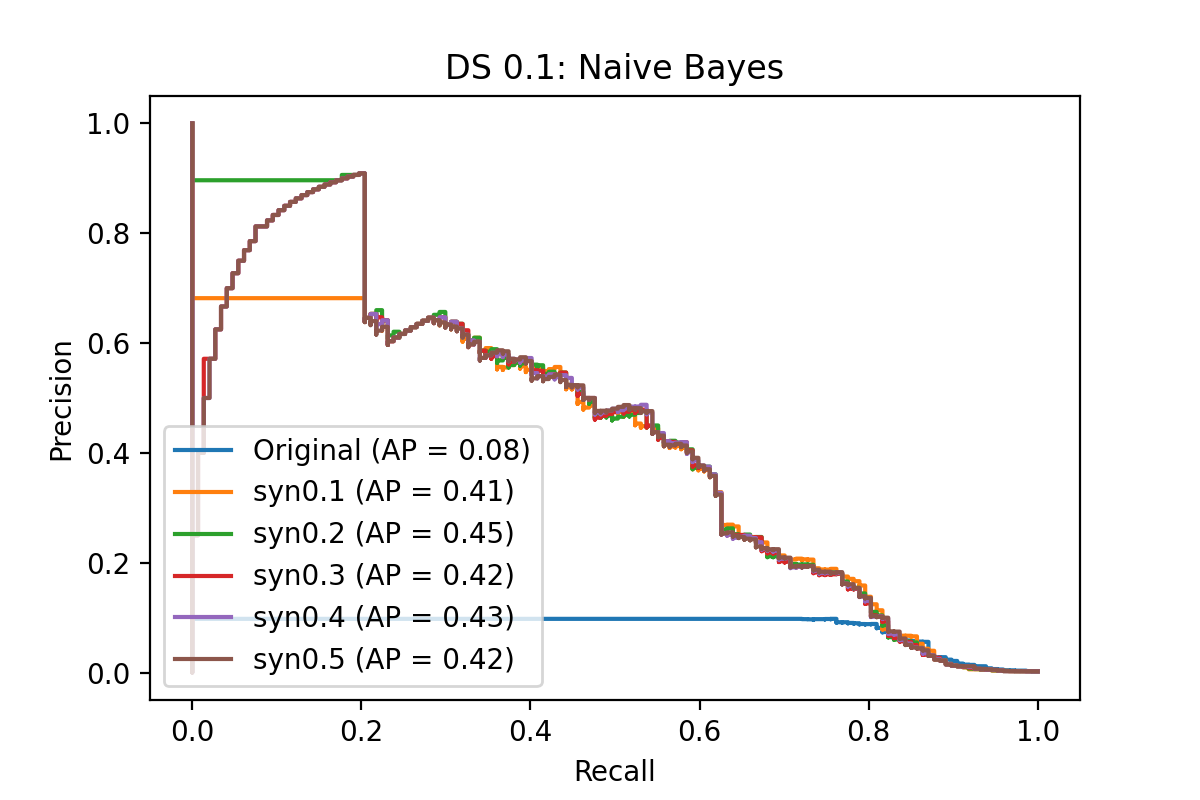}
    \includegraphics[width=.32\textwidth]{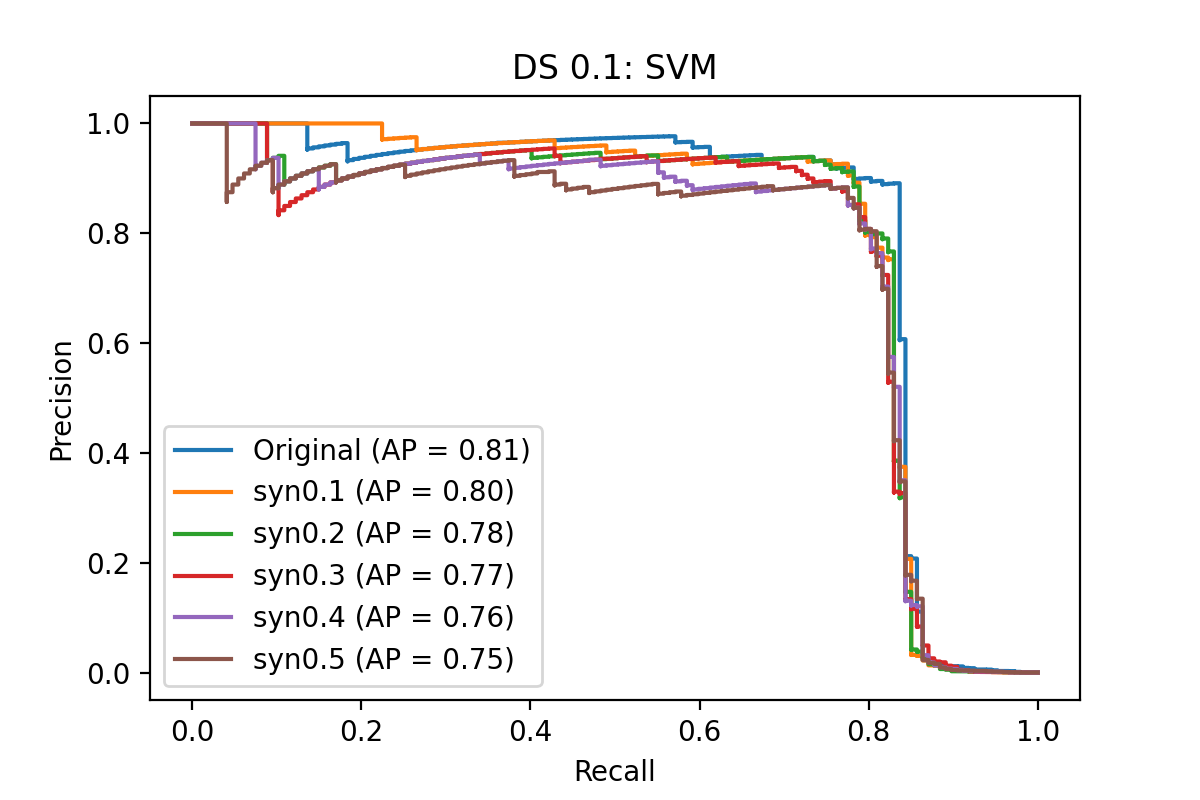}
    \includegraphics[width=.32\textwidth]{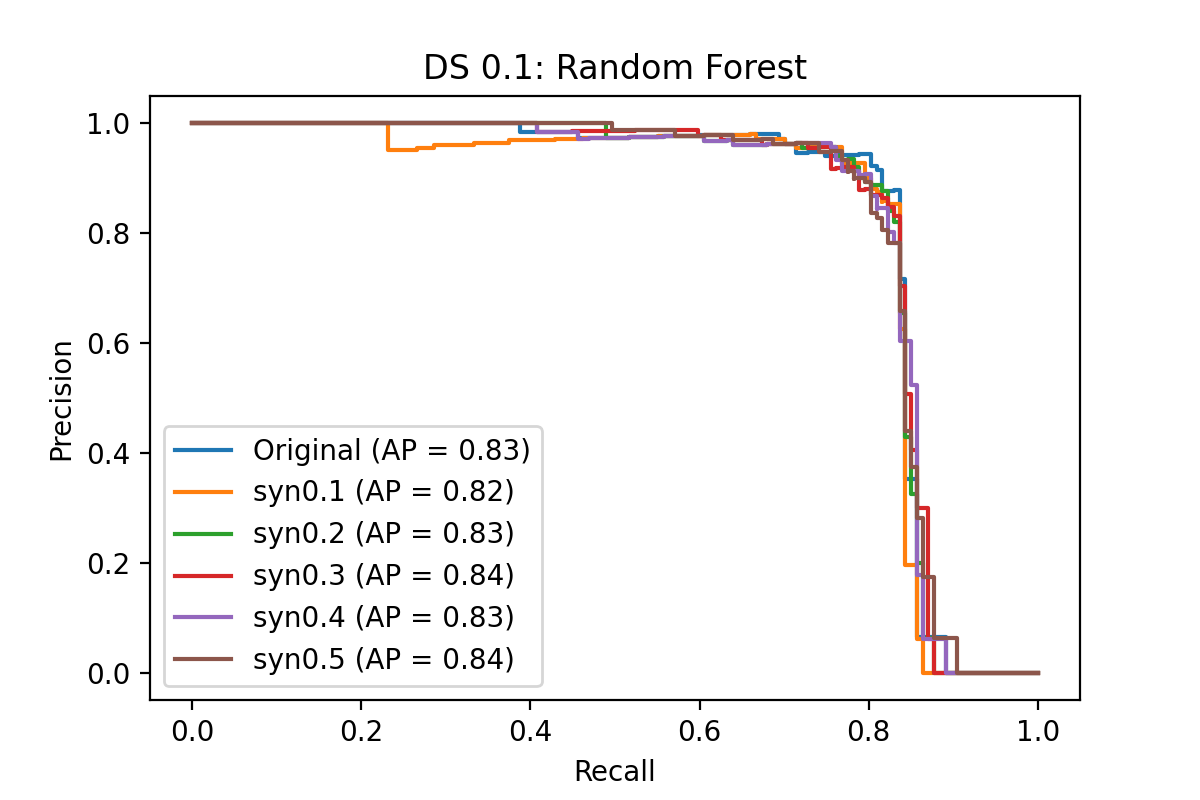}
    \includegraphics[width=.32\textwidth]{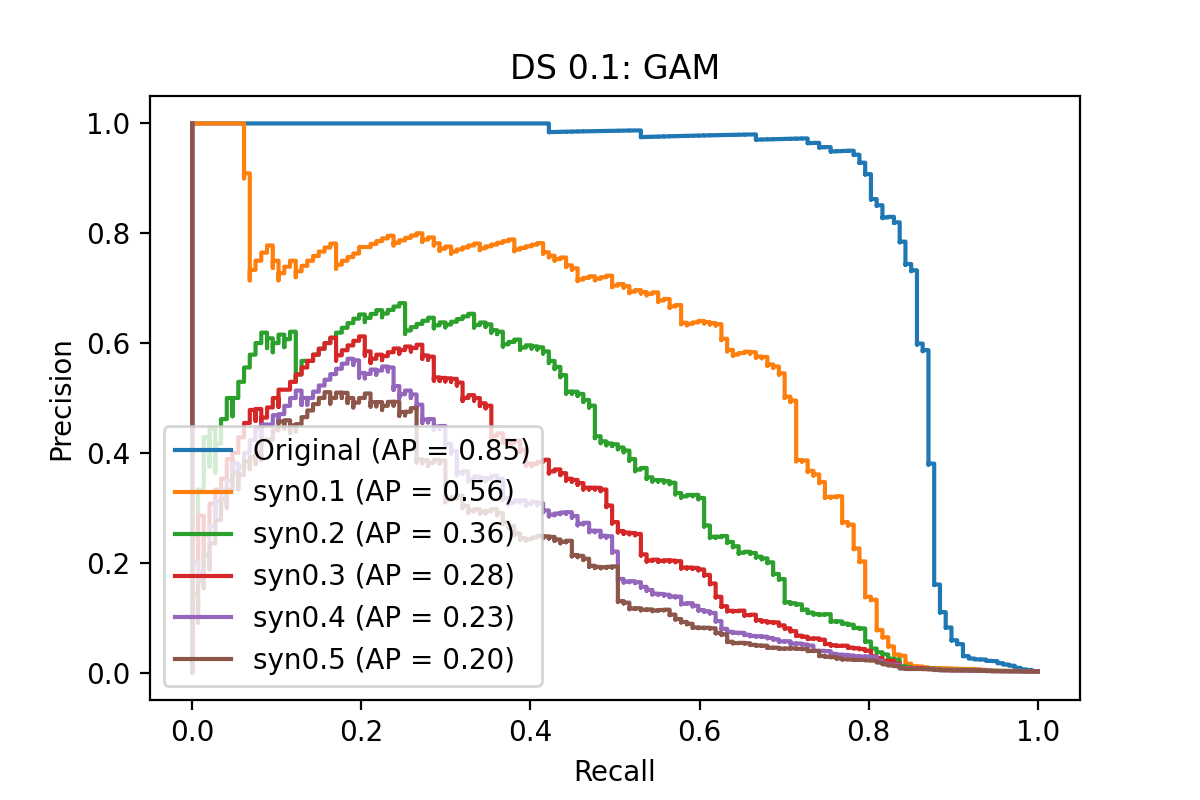}
    \includegraphics[width=.32\textwidth]{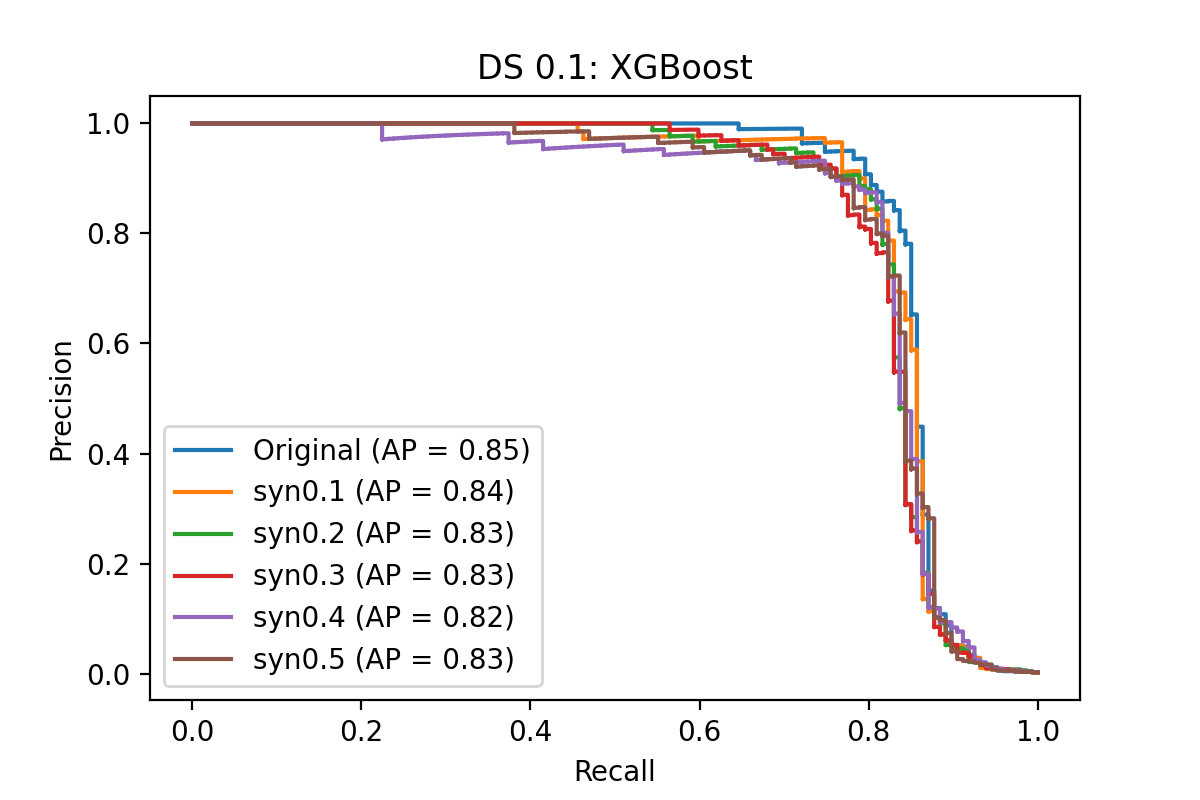}
    \includegraphics[width=.32\textwidth]{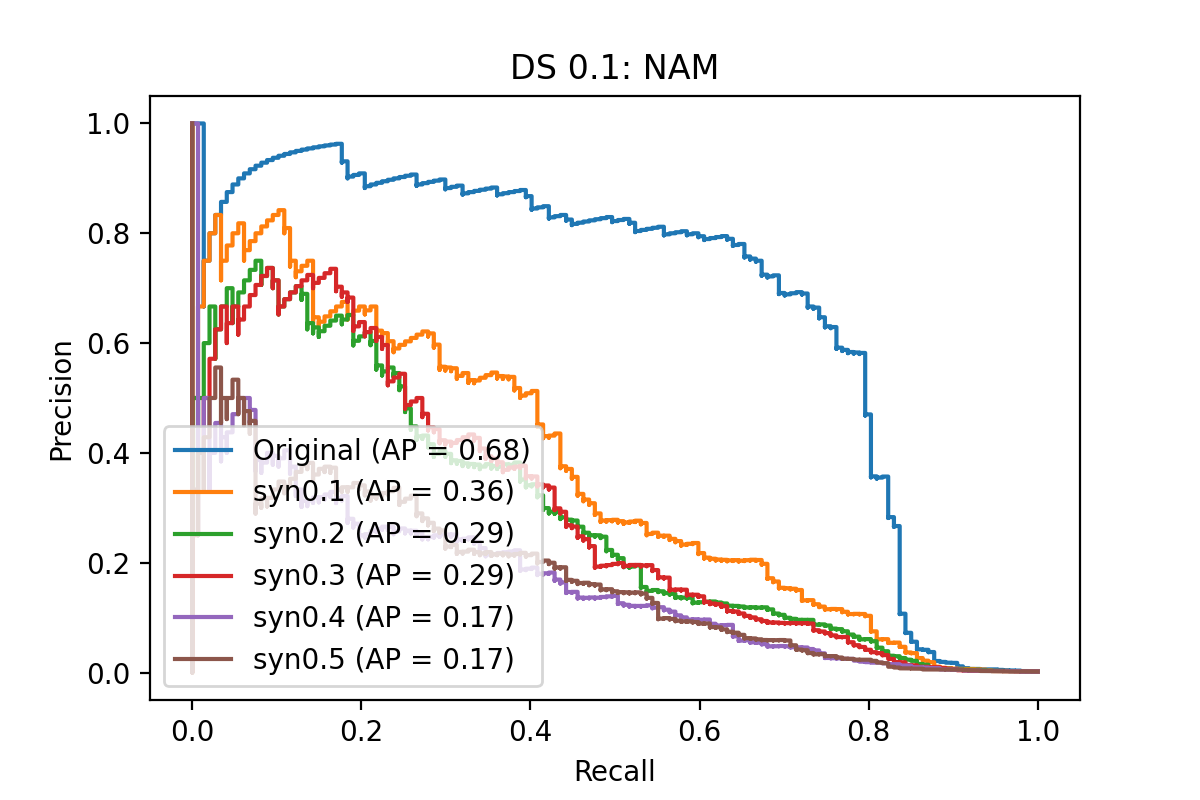}
    \caption{Precision-Recall curve for DS 0.1. DS 0.1-augmented training datasets in general damages the Precision-Recall curve across all ML classifier.}
    \label{SynAug:ds1}
\end{figure*}

\begin{itemize}
    \item Figure \ref{SynAug:ctgan} gives Synthetic Augmented Training result in Precision-Recall curve for CTGAN synthesizer.  
    \item Figure \ref{SynAug:tvae} gives Synthetic Augmented Training result in Precision-Recall curve for TVAE synthesizer.
    \item Figure \ref{SynAug:ds0} gives Synthetic Augmented Training result in Precision-Recall curve for DS0 synthesizer.
    \item Figure \ref{SynAug:ds1} gives Synthetic Augmented Training result in Precision-Recall curve for DS1 synthesizer.
\end{itemize}

\clearpage

\end{document}